\documentclass[11pt, a4paper, onecolumn, thu]{thuc3i}

\usepackage[authoryear, sort&compress, square]{natbib}
\usepackage{fontawesome5}

\usepackage{xspace}
\usepackage[table]{xcolor}

\usepackage[skip=6pt]{caption}
\definecolor{lightblue}{RGB}{0, 101, 189} %#3070B3

\usepackage{lipsum}
\usepackage{float}

\usepackage[most,skins,theorems]{tcolorbox}
\tcbset{
  takeawaysbox/.style={
    title=Takeaways,
    colback=tumblue,
    colframe=black,
    fonttitle=\bfseries\small,
    coltitle=white,
    colbacktitle=black,
    enhanced,
    attach boxed title to top left={xshift=2.5mm,yshift=-2.5mm},
    boxed title style={rounded corners, size=small, colframe=black, colback=black},
    width=\linewidth,
    arc=3.5mm
  }
}

\usepackage{graphicx}
\usepackage{algorithm}
\usepackage{algpseudocode}

\usepackage{subcaption}
\usepackage{multirow}
\usepackage{wrapfig}

\usepackage[dvipsnames]{xcolor}
\usepackage{hyperref}
\usepackage{makecell}
\usepackage{listings}
\lstnewenvironment{normalcode}{
  \lstset{
    language=Python,
    basicstyle=\ttfamily\small,
    backgroundcolor=\color{white},
    breaklines=true,
    showstringspaces=false,
    frame=none,
    xleftmargin=0.5em,
    aboveskip=1pt,
    belowskip=1pt
  }
}{}

\lstnewenvironment{diffaddcode}{
  \lstset{
    language=Python,
    basicstyle=\ttfamily\small,
    backgroundcolor=\color{green!15},
    breaklines=true,
    showstringspaces=false,
    frame=none,
    xleftmargin=0.5em,
    aboveskip=1pt,
    belowskip=1pt
  }
}{}

\lstnewenvironment{diffdelcode}{
  \lstset{
    language=Python,
    basicstyle=\ttfamily\small,
    backgroundcolor=\color{red!15},
    breaklines=true,
    showstringspaces=false,
    frame=none,
    xleftmargin=0.5em,
    aboveskip=1pt,
    belowskip=1pt
  }
}{}
\DeclareCaptionType{listing}[Listing][List of Listings]

\newcommand{\namemodule}{\textsc{StARe}}
\newcommand{\nametpo}{\textsc{StA-TPO}}
\newcommand{\nameppo}{\textsc{StA-PPO}}

\usepackage{overpic}

\definecolor{tumblue}{RGB}{0,101,189} %islexu change
\makeatletter
\AtBeginDocument{%
  \hypersetup{
    linkcolor=tumblue   % <- Figure 和 Table 会用 figblue 4 arxiv
  }%
}
\makeatother

\setheadertext{\namemodule{}-VLA}

\title{\namemodule{}-VLA: Progressive Stage-Aware Reinforcement for Fine-Tuning Vision-Language-Action Models}

\reportnumber{} %

\renewcommand{\today}{2025-12-25}

\author[*,1,3]{Feng Xu}
\author[*,1]{Guangyao Zhai}
\author[$\dagger$,2]{Xin Kong}
\author[3]{Tingzhong Fu}
\author[3]{Daniel F.N. Gordon}
\author[3]{Xueli An}
\author[1]{Benjamin Busam}

\affil[1]{\thepa{}{}}
\affil[2]{Imperial College London}
\affil[3]{Munich Research Center, Huawei Technologies}
\role[*]{Equal Contributions}
\role[\dagger]{Corresponding Author}

\resource{\faEnvelope[regular]}{EAI.Feng.Xu@tum.de, x.kong21@imperial.ac.uk}

\resource{\faGlobe}{\url{https://sites.google.com/view/starevla}}

\begin{abstract}
Recent advances in Vision-Language-Action (VLA) models, powered by large language models and reinforcement learning-based fine-tuning, have shown remarkable progress in robotic manipulation. 
Existing methods often treat long-horizon actions as linguistic sequences and apply trajectory-level optimization methods such as Trajectory-wise Preference Optimization (TPO) or Proximal Policy Optimization (PPO), leading to coarse credit assignment and unstable training. 
However, unlike language, where a unified semantic meaning is preserved despite flexible sentence order, action trajectories progress through causally chained stages with different learning difficulties. This motivates progressive stage optimization.
Thereby, we present \textbf{St}age-\textbf{A}ware \textbf{Re}inforcement (\textbf\namemodule{}), a module that decomposes a long-horizon action trajectory into semantically meaningful stages and provides dense, interpretable, and stage-aligned reinforcement signals. 
Integrating \namemodule{} into TPO and PPO, we yield Stage-Aware TPO (\textbf\nametpo{}) and Stage-Aware PPO (\textbf\nameppo{}) for offline stage-wise preference and online intra-stage interaction, respectively. 
Further building on supervised fine-tuning as initialization, we propose the \textbf{I}mitation $\to$ \textbf{P}reference$ \to$ \textbf{I}nteraction (\textbf{IPI}), a serial fine-tuning pipeline for improving action accuracy in VLA models. Experiments on SimplerEnv and ManiSkill3 demonstrate substantial gains, achieving state-of-the-art success rates of 98.0\% on SimplerEnv and 96.4\% on ManiSkill3 tasks.
\end{abstract}

\begin{document}

\maketitle

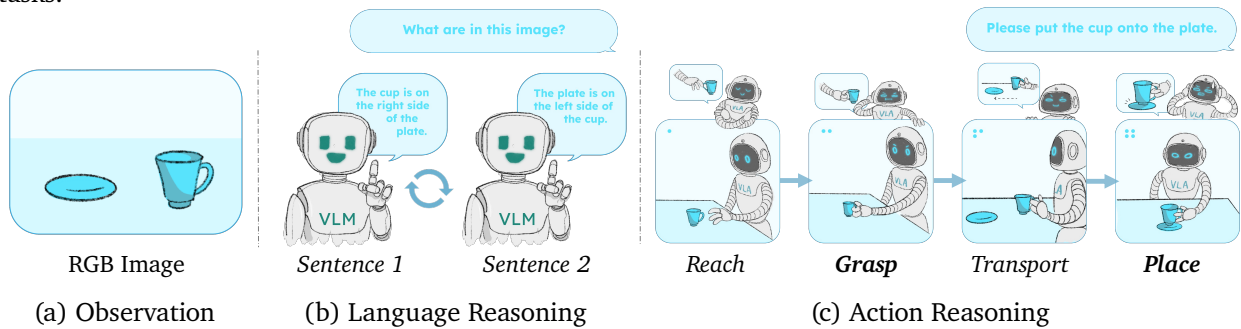
\begin{figure}[h]
  \centering
  \vspace{-3mm}
  \begin{overpic}[width=0.98\textwidth,trim=0cm 0 0 0cm,clip]{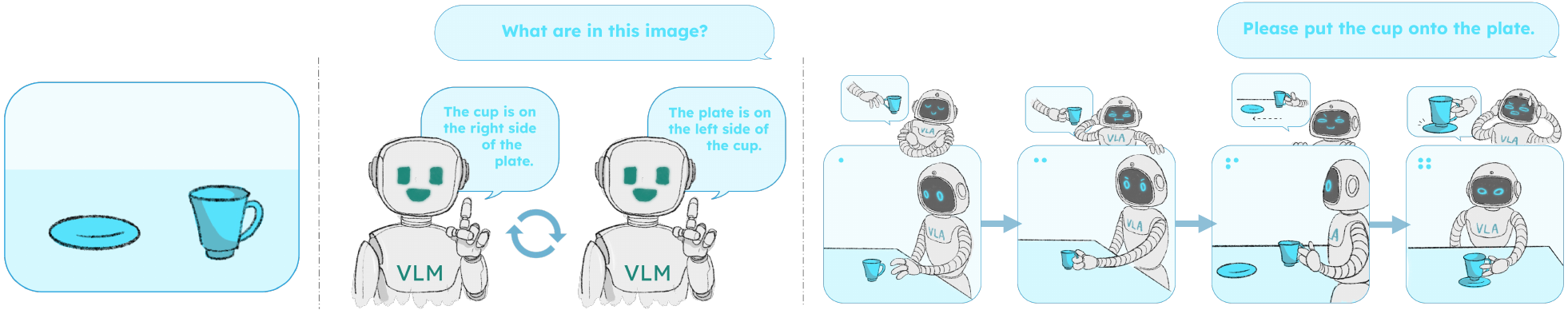}

    \put(5,-2){\color{black}{\footnotesize RGB Image}}
    \put(23.5,-2){\color{black}{\footnotesize \emph{Sentence 1}}}
    \put(38.5,-2){\color{black}{\footnotesize \emph{Sentence 2}}}
    \put(55,-2){\color{black}{\footnotesize \emph{Reach}}}
    \put(67,-2){\color{black}{\footnotesize \emph{\textbf{Grasp}}}}
    \put(78,-2){\color{black}{\footnotesize \emph{Transport}}}
    \put(92,-2){\color{black}{\footnotesize \emph{\textbf{Place}}}}
    \put(2.2,-6){\color{black}{\small (a) Observation}}
    \put(24,-6){\color{black}{\small (b) Language Reasoning}}
    \put(65,-6){\color{black}{\small (c) Action Reasoning}}
    
  \end{overpic}
  \vspace{30px} %20 to 30
    \caption{\small \textbf{Language Reasoning vs. Action Reasoning.} Given an RGB image as the observation (a), the language model (b) is asked to describe the content in the image, and produces \emph{Sentence 1} and \emph{Sentence 2}. These sentences are flexibly ordered and jointly contribute to the global meaning required to answer the question. In contrast, the VLA model (c), when instructed to place the cup onto the plate, generates an action trajectory composed of semantically meaningful stages (\emph{Reach}$\to$\emph{{Grasp}}$\to$\emph{Transport}$\to$\emph{{Place}}), which follow a strict order and vary in difficulty (with the more challenging stages shown in bold).
}
    \label{fig:teaser}
    \vspace{-5mm}
\end{figure}

\vspace{20mm}

\section{Introduction}
\label{intro}

Large-scale Vision–Language–Action (VLA) models~\citep{zitkovich2023rt,ghosh2024octo,kimopenvla,black2410pi0,intelligence2025pi_} have recently emerged as powerful generalist policies for robotic manipulation. These models unify image, language, and action modalities within a single architecture, enabling robots to interpret multimodal inputs and generate executable actions. Pretrained on massive-scale multimodal data~\citep{o2024open,walke2023bridgedata}, VLA models provide strong priors that can be efficiently adapted to diverse downstream tasks through fine-tuning, avoiding the need for retraining from scratch.

Recent development of large-scale VLA models has been rapidly driven by the success of vision–language models (VLMs) and large language models (LLMs), as their output, i.e., action trajectories and sentences, can both be represented as sequential data~\citep{zitkovich2023rt,o2024open,ghosh2024octo}. Consequently, many developed fine-tuning techniques, such as supervised fine-tuning (SFT), Reinforcement Learning from Human Feedback (RLHF) \citep{ouyang2022training}, direct preference optimization (DPO)~\citep{rafailov2023direct}, Proximal Policy Optimization (PPO) \citep{schulman2017proximal}, and Group Relative Policy Optimization (GRPO)~\citep{guo2025deepseek}, have been straightforwardly adopted for VLA models. However, directly applying these methods to fine-tune on the whole action trajectories remains cumbersome and often inefficient, as the large optimization space makes credit assignment across long-horizon trajectories highly ambiguous. Unlike language reasoning, where optimization depends on a holistic understanding of sentences without strict ordering, an action trajectory naturally decomposes into semantically distinct stages that are causally chained and vary in difficulty. For example, as in the pick-and-place task illustrated in Figure~\ref{fig:teaser}, \emph{Reach} must precede \emph{Grasp}, which in turn precedes \emph{Transport} and \emph{Place}. \emph{Reach} and \emph{Transport} are relatively easy with simple optimization objectives, while \emph{Grasp} and \emph{Place} are more challenging as they require precise geometric constraints. Overall task success hinges on correct progression through all stages. This fundamental characteristic motivates stage-aware objectives rather than monolithic trajectory-level optimization, which remains the predominant paradigm in current VLA fine-tuning.

\begin{figure}[h]
    \centering
    \includegraphics[width=1\linewidth]{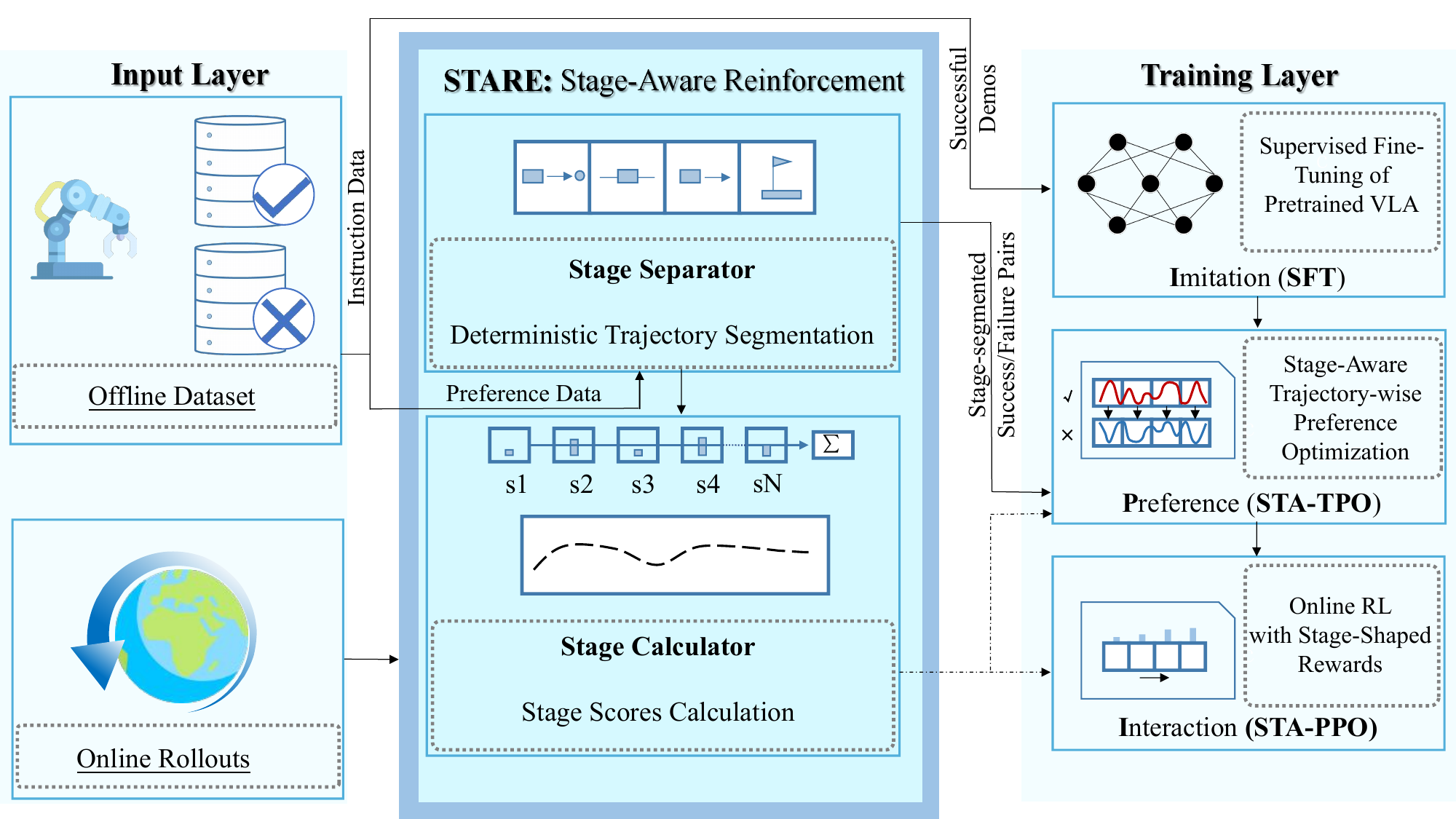}
    \caption{Overview of the \namemodule~Framework and Its Integration into the IPI Training Pipeline.}
    \vspace{-3mm}
    \label{fig:tasks}
\end{figure}

In this paper, we design \emph{\textbf{St}age-\textbf{A}ware \textbf{Re}inforcement} (\namemodule{}), a {plug-in module} that decomposes action trajectories into progressive stages with dense reward signals based on task-specific semantics. 
Given a trajectory, either in the collected data or during the model's rollout, \namemodule{} employs a stage separator to identify \emph{when} stage transitions occur, based on the translation and orientation of an end-effector. A stage calculator computes a stage cost and per-step rewards to evaluate \emph{how well} each stage is executed. In this way, \namemodule{} not only annotates stage-wise actions but also assesses partial successes and failures within a trajectory.
We leverage \namemodule{} for offline fine-tuning via \emph{Stage-Aware Trajectory-Wise Preference Optimization} (\nametpo{}), which constructs pairwise preferences at the stage level. By incorporating stage costs, \nametpo{} propagates precise gradient signals to specific action stages, enabling progressive learning and credit assignment not only between success and failure but also among varying degrees of success.
For online fine-tuning, we introduce \namemodule{} to \emph{Stage-Aware Proximal Policy Optimization} (\nameppo{}), which reshapes sparse terminal rewards to dense interaction rewards. By providing this progressive feedback, \nameppo{} stabilizes intra-stage updates, especially for complex manipulation tasks that require dense guidance. 
Conceptually, \nametpo{} and \nameppo{} are reminiscent of curriculum learning~\citep{bengio2009curriculum}, where training is organized along an ordered sequence of subtasks to ease optimization and improve generalization. However, unlike conventional curricula that progress strictly from easy to hard, our stage-aware design enforces semantic continuity across stages, ensuring that optimization respects causal dependencies in stages.

To sufficiently fine-tune a pre-trained VLA model with \nametpo{} and \nameppo{}, we integrate these two algorithms with SFT as an initialization into a serial tri-step fine-tuning pipeline, \emph{\textbf{I}mitation $\to$ \textbf{P}reference $\to$ \textbf{I}nteraction} (IPI). The IPI framework first finetunes a VLA model from expert demonstrations via SFT, then further optimizes it according to offline stage-aware preferences using \nametpo{}, and finally refines it based on stage-aware interaction in online environments using \nameppo{}. 
In contrast to existing VLA fine-tuning strategies, IPI offers two key advantages: first, it explicitly models the {multi-stage structure} of robot trajectories, enabling more precise credit assignment rather than monolithic trajectory-level optimization~\citep{zhang2024grape}. Second, compared to other methods that treat offline and online fine-tuning as disjoint processes~\citep{zhang2024grape,lu2025vlarl,chen2025conrft}, IPI unifies them under a single framework, enabling stage-wise preference alignment and intra-stage interactions. 
Extensive experiments show that IPI not only improves in-distribution success rates but also significantly enhances out-of-distribution generalization, underscoring the importance of {multi-stage reward design} in VLA fine-tuning. 

Our contributions are summarized as:
\textbf{(i)} We design \namemodule{}, a {rule-based} module that decomposes trajectories into semantically meaningful stages, enabling fine-grained supervision beyond trajectory-level signals.
\textbf{(ii)} Based on (i), we propose {stage-aware fine-tuning methods}: \nametpo{} for offline stage-wise preference alignment and \nameppo{} for online intra-stage interaction, both providing more precise credit assignment and improved sample efficiency.
\textbf{(iii)} We unify supervised fine-tuning, \nametpo{}, and \nameppo{} into {IPI}, a serial tri-step pipeline for fine-tuning VLA models, and validate it on the benchmarking frameworks SimplerEnv and ManiSkill3, showing that IPI achieves state-of-the-art success rates. %todo generalization

\newpage

\section{Related Work}

% \paragraph{Long-horizon Robotic Manipulation Tasks in RL}
% Long-horizon robotic manipulation tasks refer to operations where robots need to complete a sequence of sub-tasks over a long horizon, with frequent changes in state and environment. Applying RL to these kinds of tasks would meet many challenges brought by, such as sparse rewards, credit assignment, error accumulation, and high-dimensional state spaces. Plan-Seq-Learn~\citep{dalal2024planseqlearnlanguagemodelguided} uses language models for high-level planning and RL for low-level control, enabling an end-to-end system from visual input to complex task execution. AC$^3$~\citep{yang2025actorcriticcontinuousactionchunks} introduces an RL framework to learn continuous action chunks with intrinsic rewards to effectively tackle sparse rewards in long-horizon tasks. 
% DEMO$^3$~\citep{escoriza2025multistage} combines limited demonstrations, a world model, and stage-wise dense rewards to significantly improve sample efficiency in multi-stage visual tasks. RoboHorizon~\citep{chen2025robohorizonllmassistedmultiviewworld} uses LLMs to generate sub-goals and rewards, combined with multi-view world models and planning to achieve high success rates in long-horizon tasks. ARCH~\citep{sun2025archhierarchicalhybridlearning} trains a high-level policy to select appropriate primitive skills from a low-level primitive library to handle contact-rich assembly tasks. 
\paragraph{Long-horizon Robotic Manipulation Tasks in RL}
Long-horizon robotic manipulation involves completing a sequence of sub-tasks with frequent state and environment changes. Applying RL to such tasks is challenging due to sparse rewards, credit assignment, error accumulation, and high-dimensional state spaces. To address these issues, Plan-Seq-Learn~\citep{dalal2024planseqlearnlanguagemodelguided} leverages language models for high-level planning and RL for low-level control, enabling end-to-end execution from visual input to complex tasks. AC$^3$~\citep{yang2025actorcriticcontinuousactionchunks} learns continuous action chunks with intrinsic rewards to mitigate sparsity. DEMO$^3$~\citep{escoriza2025multistage} augments limited demonstrations with a world model and stage-wise dense rewards to improve sample efficiency. RoboHorizon~\citep{chen2025robohorizonllmassistedmultiviewworld} employs LLMs to generate sub-goals and rewards, integrated with multi-view world models and planning to achieve high success rates. ARCH~\citep{sun2025archhierarchicalhybridlearning} combines high-level policy selection with a primitive skill library to tackle contact-rich assembly.
SARM~\citep{chen2025sarmstageawarerewardmodeling} uses subtask-annotated reward modeling to filter demonstration quality, enabling robust long-horizon deformable object manipulation.
Building on these inspirations that divide goals into sub-goals, we take a further step with stage-aware reinforcement, decomposing trajectories into semantically meaningful stages. This provides denser feedback and enables progressive optimization, making RL more effective for long-horizon VLA tasks.

% \paragraph{RL Fine-tuning for LLMs}
% In the field of LLMs, RL fine-tuning has become a widely used approach for alignment and optimization. The most well-known method is Reinforcement Learning from Human Feedback (RLHF)~\citep{ouyang2022training}, where a reward model is trained using human preference data, and then algorithms such as policy gradient or PPO are trained to better align the language model’s outputs with human expectations. 
% While RLHF demonstrates impressive success, it comes with some drawbacks, such as the expense of collecting high-quality preference data, unstable training.
% To address these limitations, some work has recently been proposed.
% For example, DPO~\citep{rafailov2023direct} removes the need for a separate reward model by directly optimizing on preference comparisons, which makes the training simpler and more stable.
% Other variants like RLAIF~\citep{lee2023rlaif} and RAFT~\citep{dong2023raft}, further address the limitations of traditional RLHF.
% As a subclass of LLMs, Large Reasoning Models (LRMs)~\citep{zhang2025surveyreinforcementlearninglarge} are designed for complex multi-step and multi-stage reasoning. They face similar challenges to those in VLA long-horizon tasks, such as sparse rewards and credit assignment. DeepSeek-R1~\citep{guo2025deepseek} addresses these issues by using the GRPO algorithm, which introduces group-wise relative ranking, helping to achieve much stronger multi-step reasoning performance. This work motivates our work for VLA long-horizon tasks.
\paragraph{RL Fine-tuning for LLMs}
RL fine-tuning is a widely used approach for aligning LLMs. The most prominent method is RLHF~\citep{ouyang2022training}, where a reward model trained from human preference data guides algorithms such as policy gradient or PPO to align outputs with human expectations. While highly successful, RLHF suffers from costly data collection and unstable training. To mitigate these issues, DPO~\citep{rafailov2023direct} eliminates the reward model by directly optimizing on preference comparisons, simplifying training and improving stability. Further variants such as RLAIF~\citep{lee2023rlaif} and RAFT~\citep{dong2023raft} refine the framework.
DeepSeek-R1~\citep{guo2025deepseek} employs GRPO, which samples multiple responses per prompt and uses their relative performance within 
the group to compute advantages.
As a subclass of LLMs, Large Reasoning Models (LRMs)~\citep{zhang2025surveyreinforcementlearninglarge} utilize Chain-of-Thought (CoT) ~\citep{NEURIPS2022_9d560961} or Process Reward Model (PRM)~\citep{lightman2023let} for multi-step reasoning and face challenges akin to long-horizon VLA tasks, including sparse rewards and difficult credit assignment.  This motivates our stage-aware reinforcement approach for VLA models.

\paragraph{RL Fine-tuning for VLAs}
Recent studies explore RL as a fine-tuning paradigm for VLA models. GRAPE~\citep{zhang2024grape} adapts DPO~\citep{rafailov2023direct} to trajectory-level preferences to propose TPO, while ConRFT~\citep{chen2025conrft} alternates RL and SFT in real-world settings. ReinboT~\citep{zhang2025reinbot} designs dense rewards, and \cite{guo2025improving} propose an iterative SFT–RL pipeline to reduce instability and cost. RIPT-VLA~\citep{tan2025interactive} applies RLOO~\citep{ahmadian2024back} for online training, RL4VLA~\citep{liu2025rlbringvlageneralization} studies RL-driven generalization, VLA-RL~\citep{lu2025vlarl} applies PPO, and RFTF~\citep{shu2025rftf} introduces value models for dense reward estimation. SimpleVLA-RL~\citep{li2025simplevla} extends veRL to VLA models with GRPO-based online RL, demonstrating significant improvements in data efficiency, long-horizon task performance, and generalization across spatial, object, and task distributions.
RLinf~\citep{zang2025rlinf} provides a scalable and unified pipeline for VLA RL—combining rendering, inference, and training—to boost efficiency and performance.
$\pi_{\texttt{RL}}$~\citep{chen2025pi_} applied online RL fine-tuning for flow-based VLAs.
{$\pi^{*}_{0.6}$}~\citep{intelligence2025pi06vlalearnsexperience} uses RECAP~\citep{intelligence2025pi06vlalearnsexperience} to fine-tune VLAs with advantage-conditioned policies, learning from autonomous experience and expert corrections. GR-RL~\citep{li2025gr} combines offline data filtering with distributional critics and online latent-space RL for high-precision dexterous manipulation.
Despite their promise, these methods typically optimize at the trajectory level, suffering from sparse rewards, coarse credit assignment, and difficult exploration in long-horizon manipulation. In contrast, our stage-aware RL decomposes trajectories into semantically meaningful stages and assigns stage-level rewards, providing denser, interpretable feedback and enabling progressive optimization for complex robotic tasks.

% SimpleVLA-RL~\citep{li2025simplevla} extends veRL to VLA models with GRPO-based online RL, demonstrating significant improvements in data efficiency, long-horizon task performance, and generalization across spatial, object, and task distributions.

% $\pi_{\texttt{RL}}$~\citep{chen2025pitextttrlonlinerlfinetuning} applied online RL fine-tuning for flow-based VLAs, introducing Flow-Noise and Flow-SDE to address the intractable log-likelihood estimation problem in flow matching.

% {$\pi^{*}_{0.6}$}~\citep{intelligence2025pi06vlalearnsexperience} uses RECAP to fine-tune VLAs with advantage-conditioned policies, learning from autonomous experience and expert corrections.

\section{Preliminary}

\subsection{Problem Formulation}
We consider a language-conditioned POMDP problem defined by the tuple $\{\mathcal{S}, \mathcal{A}, \mathcal{T}, \mathcal{L}, \mathcal{R}, \gamma\}$, where $\mathcal{S}$ is the state space, $\mathcal{A}$ is the action space, $\mathcal{T}: \mathcal{S} \times \mathcal{A} \rightarrow \mathcal{S}$ is the dynamic function, $\mathcal{L}$ is the space of language instruction, $\mathcal{R}: \mathcal{S} \times \mathcal{L} \rightarrow \mathbb{R}$ is the reward function, and $\gamma$ is a scale factor with $0<\gamma<1$. 
The goal of a VLA model is to find a policy $\pi_\theta: \mathcal{S} \times \mathcal{L} \rightarrow \mathcal{A}$, which generates action trajectories maximizing the expected accumulated reward, or return for each task $l$, i.e.\ $\mathcal{R}(\pi,l) = \mathbb{E}_{a \sim \pi} [\sum_t \gamma^tr_t]$.

Fine-tuning a VLA model adapts a pre-trained $\pi_\theta$ to new tasks so that the resulting policy $\pi_{\theta'}$ maximizes expected return under the POMDP. This can be done through imitation for aligning with expert demonstrations, preference for refining trajectories via learned comparisons, or reinforcement optimizing long-term rewards.

% \newpage

\subsection{Trajectory-Wise Preference Optimization (TPO)}
Direct Preference Optimization (DPO)~\citep{rafailov2023direct} is a recent fine-tuning technique for large language models that directly aligns a policy with preference data, bypassing explicit reward modeling. Extending this idea to fine-tuning VLA models yields TPO~\citep{zhang2024grape}: the outputs are action trajectories $\tau = \{(s_t,a_t)\}_{t=1}^T$ rather than text sequences. TPO treats each trajectory as a single sequence and learns from pairwise comparisons of successful and failed trajectories $(\tau^+,\tau^-)$ generated under the same instruction. The policy is updated to prefer $\tau^+$ over $\tau^-$ by minimizing
\begin{subequations}\label{eq:tpo-loss}
\begin{align}
{L}_{\text{TPO}}(\theta) 
&= -\mathbb{E}_{(\tau^+, \tau^-)} \Big[ 
\log \sigma\big( \beta(q(\tau^+) - q(\tau^-)) \big)
\Big], \label{eq:stage-policy-loss} \\
q(\tau) 
&=  \frac{1}{T}\sum_{t=1}^{T} 
\Big(\log \pi_{\theta'}(a_t|s_t) - \log \pi_{\theta}(a_t|s_t)\Big).
\label{eq:tpo-preference}
\end{align}
\end{subequations}
where $\sigma(\cdot)$ is the sigmoid function, 
$\beta$ controls the strength of preference alignment, 
$s_t \in \mathcal{S}$ and $a_t \in \mathcal{A}$ denote the environment state and action at timestep $t$, 
and $q(\cdot)$ measures the normalized log-likelihood ratio of a trajectory under policy $\pi_{\theta'}$ relative to $\pi_{\theta}$.
${L}_{\text{TPO}}$ is minimized when the model increases $q(\tau^+)$ relative to $q(\tau^-)$, 
i.e., when the likelihood of successful trajectories exceeds
% that of failed ones.
failed ones.

While TPO provides a direct mechanism to apply preference learning to long-horizon control, it suffers from credit assignment ambiguity: preferences are assigned to full trajectories, 
making it difficult to determine which specific stage contributed to the preference signal. 
Moreover, such a binary preference limits optimization to coarse distinctions between successful and failed rollouts, without capturing relative quality among partially successful trajectories. These limitations motivate (\nametpo{}), which decomposes trajectories into stages and aligns hierarchical preferences at the stage level, enabling finer-grained optimization.

\subsection{Proximal Policy Optimization (PPO)}
\label{ppo}
PPO~\citep{schulman2017proximal} is one of the most widely used online reinforcement learning algorithms, known for its balance of sample efficiency and training stability. PPO improves policy gradient methods by introducing a clipped surrogate objective that prevents excessively large policy updates, thereby stabilizing training. Given an old policy $\pi_{\theta}$, the clipped objective is
\begin{equation}
\label{eq:ppo}
{L}_{\text{PPO}}(\theta) = 
\mathbb{E}_t \Big[
\min \big(
p_t(\theta) \operatorname{GAE}(r_t), \;
\text{clip}(p_t(\theta), 1-\epsilon, 1+\epsilon)\operatorname{GAE}(r_t)
\big)
\Big],
\end{equation}
where $p_t(\theta) = {\pi_{\theta'}(a_t|s_t)}/{\pi_{\theta}(a_t|s_t)}$ is the likelihood ratio between the new and old policies, and $\epsilon$ is a clipping parameter. $\operatorname{GAE}(\cdot)$ is generalized advantage estimator~\citep{schulman2018highdimensionalcontinuouscontrolusing} that estimate the advantage value based on rewards $r_t$. 

In the context of fine-tuning VLA models, PPO is commonly used to fine-tune policies with sparse $r_t$, but such signals often limit sample efficiency and provide insufficient guidance for complex, long-horizon tasks. This motivates \nameppo{}, which integrates stage-aware reward shaping to transform sparse terminal rewards into dense progressive signals for more efficient fine-tuning.

% \input{sections/3.method}
% =================================
% Methodology (STage-WISE)
% =================================

\section{Method}
We begin by introducing \emph{Stage-Aware Reinforcement} (\namemodule{}), which decomposes long-horizon action trajectories into semantically meaningful stages, each equipped with stage-wise costs and intra-stage rewards.
Building on this foundation, we develop offline and online learning algorithms for progressive stage-aware fine-tuning: \emph{Stage-Aware Trajectory Preference Optimization} (\nametpo{}) and \emph{Stage-Aware Proximal Policy Optimization} (\nameppo{}). 
Finally, we integrate \nametpo{} and \nameppo{} with supervised fine-tuning (SFT) into a serial pipeline, \emph{Imitation$ \to$ Preference $\to$ Interaction} (IPI), to achieve sufficient fine-tuning of VLA models.

% STARE
\subsection{Stage-Aware Reinforcement (\namemodule{})}
\label{sec:stare}

We propose \namemodule{}, a module that decomposes long-horizon action trajectories into semantically meaningful stages defined by task-specific rules. \namemodule{} consists of two components: (i) a \emph{stage separator}, which determines \emph{when} stage transitions occur by detecting task-relevant events, and (ii) a \emph{stage calculator}, which evaluates \emph{how well} each stage is executed using stage-wise costs and dense intra-stage rewards.

\paragraph{Stage Separator}
Stage boundaries are determined by semantically meaningful manipulation events 
rather than arbitrary temporal cuts. 
Given the whole action trajectory $\tau$, we intend to divide it into $K$ stages by defining semantic boundaries and assigning each global timestep $t$ a stage label $k \in \{1,\dots,K\}$.
Following an event-driven rule, the entry condition of stage $k$ coincides with the terminal condition of stage $k-1$, ensuring progressive continuity across stages. For instance, a pick-and-place task can be separated into four stages: \emph{Reach} $\rightarrow$ \emph{Grasp} $\rightarrow$ \emph{Transport} $\rightarrow$ \emph{Place}.

Stage segmentation thus reduces to detecting the onset of each stage based on geometric constraints, defined by thresholds $\delta_k$ on the translation and orientation signals of the end-effector. These thresholds set binary environment flags (e.g., grasped, on-target). For example: a \emph{Reach} $\rightarrow$ \emph{Grasp} transition occurs when the end-effector contacts the object; \emph{Grasp} $\rightarrow$ \emph{Transport} occurs when the grasped object is lifted above a small height threshold; \emph{Transport} $\rightarrow$ \emph{Place} occurs when the object is within a distance margin of the goal position; and \emph{Place} $\rightarrow$ \emph{Success} occurs when the object is released and remains stably in the goal region (see  Supplementary Material~\ref{app:stagewisecostandpotential} for other segmentation examples). Thereby, we group steps with the same stage label into the $k$-th trajectory segment
$\tau^{(k)} = \{(s_t,a_t)\mid g(t) = k\}_{t=1}^{T_k}$,
where $s_t \in \mathcal{S}$, $a_t \in \mathcal{A}$, 
and $T_k$ is the number of timesteps assigned to stage $k$. 
Here, $g: \mathbb{N} \to \{1, \dots, K\}$ is a stage assignment function mapping each timestep $t$ 
to its corresponding stage index $k$. 
The full trajectory can then be expressed as the stage-wise decomposition
\(\tau \mapsto \{\tau^{(i)}\}_{i=1}^{K}\).

\paragraph{Stage Calculator}
Given the stage segments produced by the stage separator, the stage calculator computes both stage-wise costs and intra-stage dense rewards by measuring the relation between the end-effector and relevant targets in the environment. The specific forms of cost and reward depend on the goal of each stage. We illustrate with \emph{Reach} as the $k$-th stage:

\quad (i) \emph{Stage cost aggregation.}
We define the cost function $\ell_k(\cdot)$ as the mean Euclidean distance over $T_k$ between the end-effector and the target object from start to the end of \emph{Reach}:
\begin{equation}
\ell_k(\tau^{(k)}) = \frac{1}{T_k}\sum_{t=1}^{T_k} 
\|x_{\mathrm{ee}}(t) - x_{\mathrm{obj}}(t)\|_2,
\end{equation}
where $x_{\mathrm{ee}}(t) \in \mathbb{R}^3$ denotes the Cartesian position of the end-effector 
at time step $t$, and $x_{\mathrm{obj}}(t) \in \mathbb{R}^3$ is the target position of the object. 
By definition, $\ell_k$ is a non-negative value measuring the deviation from the target: 
the better the $\tau^{(k)}$, the smaller the $\ell_k$. 
Detailed cost functions for other stage categories are provided in the Supplementary Material~\ref{app:stagewisecostandpotential}.

\quad (ii) \emph{Intra-stage reward shaping.}
To provide dense guidance, we adopt potential-based reward shaping~\citep{kim2024stagewiserewardshapingacrobatic,ng1999policy}. For active stage $k$, we define a per-timestep potential $\Phi_{k_t}$ that captures the normalized progress of state $s_t$. Specifically, for \emph{Reach}, we use:
\begin{equation}
\Phi_{k_t}(s_t)
= \sigma\!\Big(1 - \tfrac{\|x_{\mathrm{ee}(t)} - x_{\mathrm{obj(t)}}\|}{d_{k}}\Big),
\end{equation}
where $\Phi_{k_t}(s_t) \in [0,1]$, $d_k$ is a normalization length scale, and $\sigma(\cdot)$ is a sigmoid function. This provides smooth shaping rewards that encourage the end-effector to progressively reach the target (see detailed potential functions for other stages in the Supplementary Material~\ref{app:stagewisecostandpotential}).
Based on $\Phi_{k_t}$, the shaped reward $r'_t$ augments the sparse reward $r_t$ as:
\begin{equation}
\label{eq:star-pbrs}
r'_t \;=\; r_t \;+\; \gamma \,\Phi_{k_{t+1}}(s_{t+1}) \;-\; \Phi_{k_t}(s_t).
\end{equation}

% STA-TPO
\subsection{From \namemodule{} to \nametpo{}}
\label{sec:satpo}

Unlike standard TPO~\citep{zhang2024grape}, which aggregates preferences only at the level of entire trajectories, \nametpo{} leverages \namemodule{} to segment trajectories into progressive stages and perform stage-wise preference alignment. A detailed algorithm is shown in Algorithm~\ref{alg:sta-tpo}. A pair comparison of stage samples $(\tau^{(k)+}, \tau^{(k)-})$ exists only when the previous stage $\tau^{(k-1)}$ has been successfully completed, ensuring progressive consistency across stages. In addition, the stage cost $\ell_k(\tau)$ is incorporated as a penalty term in \eqref{eq:tpo-preference}, transforming $q$ into $\hat{q}$:
\begin{equation}
\hat{q}(\tau^{(k)}) =
{q}(\tau^{(k)})
- \lambda \ell_k(\tau^{(k)}),
\label{eq:stage-policy}
\end{equation}
where $\lambda$ is the penalty weight. The original objective $\mathcal{L}_{\text{TPO}}$ in \eqref{eq:stage-policy-loss} thereby extends to $\mathcal{L}_{\nametpo{}}$.
Compared to \eqref{eq:tpo-loss}, which optimizes the model only with binary trajectory-level preferences (success vs.\ failure), $\hat{q}$ introduces a hierarchical signal to $\mathcal{L}_{\nametpo{}}$. Even among successful stages $\tau^{(k)+}$ across different trajectories, those with lower penalties $\ell_k(\tau^{(k)})$ yield higher $\hat{q}$, while less optimal stages receive lower $\hat{q}$. This design enables credit assignment not only between success and failure but also among varying degrees of success, thereby providing finer-grained supervision for learning optimal behaviors.

\begin{algorithm}[h]
\caption{\nametpo{} (offline)}
\label{alg:sta-tpo}
\begin{algorithmic}[1]
\Require Preference pairs $\{(\tau^+,\tau^-)\}$ under same instruction; reference policy $\pi_\theta$; \namemodule{}; stage penalty weight $\lambda$; temperature $\beta$; learning rate $\eta$.
\Ensure Updated policy $\pi_{\theta'}$.
\While{not converged}
  \State Sample minibatch $\mathcal{B}=\{(\tau^+,\tau^-)\}$.
  \ForAll{$\tau \in \{\tau^+,\tau^-\}$ in $\mathcal{B}$}
    \State \textbf{Stage segmentation \& costs:} $\{\tau^{(k)},\ell_k(\tau^{(k)})\}_{k=1}^{K} \gets \namemodule{}(\tau)$.
    \State \textbf{Stage scores:} For each $k$, first compute
      \[
      q(\tau^{(k)}) \gets \frac{1}{T_k}\!\sum_{t\in T_k}\!\Big(\log \pi_{\theta'}(a_t|s_t) - \log \pi_{\theta}(a_t|s_t)\Big).
      \tag*{(cf.\ ~\eqref{eq:tpo-preference})}
      \]
    \State Then add stage costs: $\hat{q}(\tau^{(k)}) \gets q(\tau^{(k)}) - \lambda\,\ell_k(\tau^{(k)})$.
  \EndFor
  \State \textbf{Preference loss:} Compute
  \[
  {L}_{\nametpo{}} \gets -\frac{1}{|\mathcal{B}|}\sum_{(\tau^+,\tau^-)\in\mathcal{B}}\frac{1}{K}
  \sum_{k=1}^{K}\log \sigma\!\Big(\beta\big(\hat{q}(\tau^{(k)+})-\hat{q}(\tau^{(k)-})\big)\Big).
  \tag*{(cf.\ ~\eqref{eq:stage-policy-loss})}
  \]
  
  \State \textbf{Policy update:} $\theta' \gets \theta' - \eta\,\nabla_{\theta'}{L}_{\nametpo{}}$.
\EndWhile
\end{algorithmic}
\end{algorithm}

% STA-PPO
\subsection{From \namemodule{} to \nameppo{}}
\label{sec:sappo}

For online RL fine-tuning, we integrate \namemodule{} directly into rollouts. The stage separator determines the stage transition online. At each time step within the stage, the {stage calculator} produces shaped reward $r'_t$, turning ${L}_{\text{PPO}}$ in \eqref{eq:ppo} into ${L}_{\nameppo{}}$.
Finally, policy parameters $\theta$ are updated by minimizing ${L}_{\nameppo{}}$.
By replacing $r_t$ with $r'_t$, \nameppo{} provides denser, stage-aligned feedback that accelerates policy learning in long-horizon, sparse-reward tasks. A detailed algorithm is shown in Algorithm~\ref{alg:sta-ppo}.

\begin{algorithm}[h]
\caption{\nameppo{} (online)}
\label{alg:sta-ppo}
\begin{algorithmic}[1]
\Require Simulation $\operatorname{Env}$; Behavior policy $\pi_\theta$; \namemodule{}; horizon $T$; PPO epochs $E$; discount $\gamma$; GAE parameter; clip $\epsilon$; step size $\eta$.
\Ensure Updated policy $\pi_{\theta'}$.
\While{not converged}
  \State \textbf{Rollout:} collect $\{(s_t,a_t,r_t,\log\pi_\theta(a_t|s_t))\}_{t=0}^{T-1}$ in $\operatorname{Env}$.
  \State \textbf{Online stage labels \& potentials:}
  \For{$t=0$ to $T-1$}
    \State Detect current stage $k=g(t)$ via stage separator in \namemodule{} (event rules).
    \State Compute potential $\Phi_{k_t}(s_t)$ by the stage calculator in \namemodule{}.
  \EndFor
  \State \textbf{Shaped rewards:}
  \For{$t=0$ to $T-1$}
    \State $r'_t \gets r_t + \gamma\,\Phi_{k_{t+1}}(s_{t+1}) - \Phi_{k_t}(s_t)$ \Comment{Potential-based shaping}
  \EndFor
  \For{$e=1$ to $E$} \Comment{PPO updates}
    \State Compute ratio $p_t(\theta') = \exp(\log\pi_{\theta'}(a_t|s_t) - \log\pi_\theta(a_t|s_t))$.
    \State \textbf{Interaction loss:}
    \[
    {L}_{\nameppo{}} \gets\mathbb{E}_t\!\left[\min\big(p_t(\theta')\operatorname{GAE}(r'_t),\;\mathrm{clip}(p_t(\theta'),1-\epsilon,1+\epsilon)\operatorname{GAE}(r'_t)\big)\right].
    \tag*{(cf.\ ~\eqref{eq:ppo})}
    \]
    \State \textbf{Update:} $\theta' \gets \theta' + \eta\,\nabla_{\theta'}{L}_{\nameppo{}}$.
  \EndFor
\EndWhile
\end{algorithmic}
\end{algorithm}

% \input{sections/algorithm}

% IPE
\subsection{\nametpo{} And \nameppo{} for Serial Fine-tuning}
% This raises a natural question: \emph{why not combine them into a unified framework that benefits from both?}  
Existing works often apply offline preference-based optimization~\citep{zhang2024grape} and online RL fine-tuning~\citep{liu2025rlbringvlageneralization, li2025simplevla} separately. Besides, while we are now able to jointly address offline preference alignment and online reinforcement learning by \nametpo{} and \nameppo{}, a complete fine-tuning framework for VLA models must also incorporate {imitation learning} to initialize a strong policy prior.  

Thereby, we propose {Imitation $\to$ Preference $\to$ Interaction (IPI)}, a three-step fine-tuning pipeline. We first warm up the policy safely from demonstrations by SFT. Then we apply \nametpo{} to offline refine the policy. Finally, we apply \nameppo{} to further enhance robustness through online exploration.  
Thereby, IPI integrates supervised, preference-based, and exploration signals into a coherent progression, yielding more sample-efficient and more robust fine-tuning of VLA models.

\section{Experiments}
\label{sec:experiments}
\begin{figure}[h]
    \centering
    \includegraphics[width=0.7\linewidth]{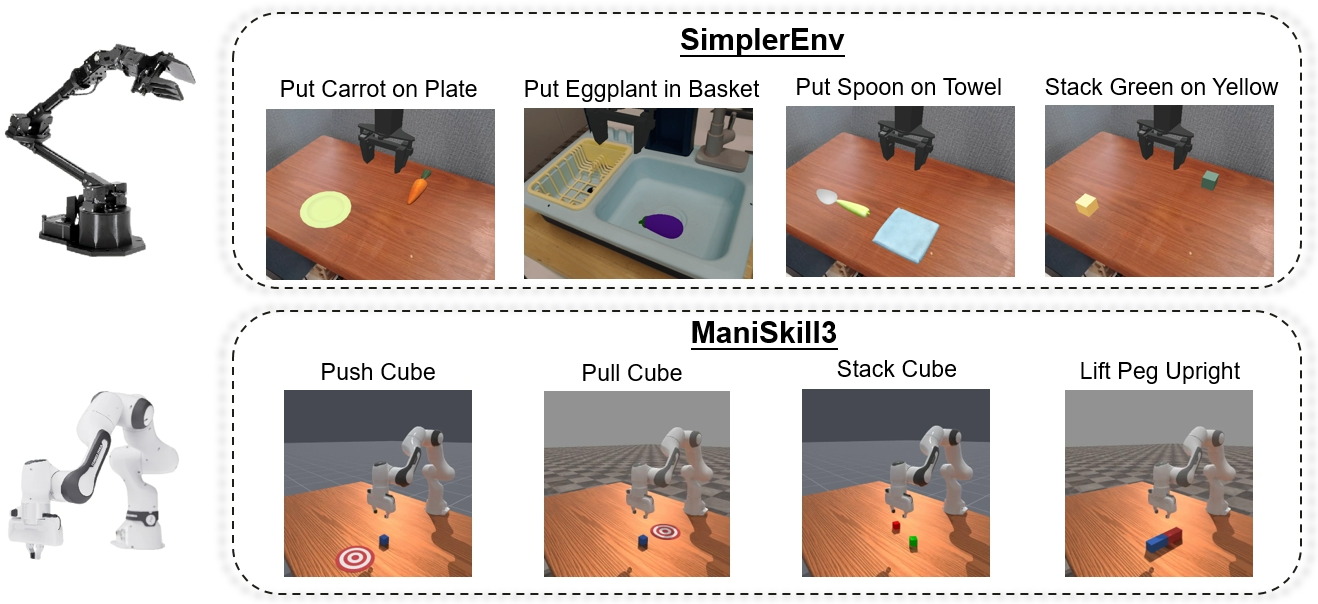}
    \caption{Two simulated benchmarks. We show experiment setups and example tasks involved.}
    \vspace{-3mm}
    \label{fig:tasks}
\end{figure}
\paragraph{Benchmarks \& Baselines.}
We evaluate our approach on two families of robotic manipulation environments, as shown in Figure~\ref{fig:tasks}. 
The first is \textbf{SimplerEnv}~\citep{li2024evaluating} with the WidowX arm, where we focus on the four canonical single-object tasks in the \textbf{SimplerEnv-WidowX} split. 
The second is \textbf{ManiSkill3}~\citep{taomaniskill3} with the Franka robot~\citep{10693652}, including \emph{StackCube} and three contact-rich tasks (\emph{PushCube}, \emph{PullCube}, and \emph{LiftPegUpright}) to validate generality beyond pick-and-place and assess performance under challenging non-trivial manipulation. 
We compare against widely used VLA baselines (RT-1-X, Octo-Base/Small, RoboVLM, SpatialVLA), the strong offline preference fine-tuning method GRAPE
~\citep{zhang2024grape}, and the RL fine-tuning baselines RL4VLA~\citep{liu2025rlbringvlageneralization} and $\pi_{\texttt{RL}}$~\citep{chen2025pi_}. 
For fairness, all methods fine-tune the {OpenVLA-7B}~\citep{kimopenvla} and {pi0.5\_base}~\citep{intelligence2025pi_} backbone, and we additionally evaluate our proposed \nametpo{}, \nameppo{}, and the full \textbf{IPI}. We report average success over 300 evaluation episodes per method and setting.
Unless otherwise stated, hyperparameters are shared across methods when applicable (detailed in Appendix~\ref{app:training}).

\begin{table*}[t]
    \centering
    \caption{\textbf{Evaluation on SimplerEnv with WidowX Robot tasks.} 
    We report the final success rate and grasp success rate shown in parentheses (Success \% (\text{\scriptsize Grasp \%})). Our method, IPI, uses RL fine-tuning after an initial SFT phase. The `{\color{green!60!black}(+X\%)}' indicates the improvement over a relevant baseline.}
    \label{tab:simplerenv_windowx}
    \setlength{\tabcolsep}{4pt}
    \resizebox{0.95\textwidth}{!}{
    \begin{tabularx}{\textwidth}{>{\raggedright\arraybackslash}X cccc c}
        \toprule
        \multirow{2}{*}{\textbf{Methods}} & \multicolumn{4}{c}{\textbf{Robotic Task}} & \multirow{2}{*}{\makecell{\textbf{Avg.} \\ \textbf{Success} \\ \textbf{Rate (\%)}}} \\
        \cmidrule(lr){2-5}
        & \makecell{Put Spoon \\ on Towel} & \makecell{Put Carrot \\ on Plate} & \makecell{Stack Green \\ on Yellow} & \makecell{Put Eggplant \\ in Basket} & \\
        \midrule
        \multicolumn{6}{l}{\textit{Other Methods}} \\
        RT-1-X \citep{Brohan-RSS-23} & 0.0 ({\scriptsize 16.7}) & 4.2 ({\scriptsize 20.8}) & 0.0 ({\scriptsize 8.3}) & 0.0 ({\scriptsize 0.0}) & 1.1 \\
        Octo-Base \citep{ghosh2024octo} & 12.5 ({\scriptsize 34.7}) & 8.3 ({\scriptsize 52.8}) & 0.0 ({\scriptsize 31.9}) & 43.1 ({\scriptsize 66.7}) & 16.0 \\
        Octo-Small \citep{ghosh2024octo} & 47.2 ({\scriptsize 77.8}) & 9.7 ({\scriptsize 27.8}) & 4.2 ({\scriptsize 40.3}) & 56.9 ({\scriptsize 87.5}) & 30.0 \\
        RoboVLM \citep{liu2025towards} & 20.8 ({\scriptsize 37.5}) & 25.0 ({\scriptsize 33.3}) & 8.3 ({\scriptsize 8.3}) & 0.0 ({\scriptsize 0.0}) & 13.5 \\
        SpatialVLA \citep{qu2025spatialvla} & 20.8 ({\scriptsize 25.0}) & 20.8 ({\scriptsize 41.7}) & 25.0 ({\scriptsize 58.3}) & 70.8 ({\scriptsize 79.2}) & 34.4 \\
        SOFAR \citep{qi2025sofar} & 58.3 ({\scriptsize 62.5}) & 66.7 ({\scriptsize 75.0}) & 70.8 ({\scriptsize 91.7}) & 37.5 ({\scriptsize 66.7}) & 58.3 \\
        \midrule
        \multicolumn{6}{l}{\textit{OpenVLA-7B Based Methods}} \\
        SFT & 43.7({\scriptsize 70.3}) & 52.7({\scriptsize 74.7}) & 21.3({\scriptsize 59.0}) & 49.0({\scriptsize 67.3}) & 41.7 \\
        GRAPE \citep{zhang2024grape} & 44.3({\scriptsize 72.0}) & 55.0({\scriptsize 85.3}) & 22.7({\scriptsize 53.3}) & 53.7({\scriptsize 78.7}) & 43.9 \\
        \rowcolor{gray!5} 
        SFT $\rightarrow$ \nametpo  (\textcolor{tumblue}{\textbf{\namemodule}})  & 51.0({\scriptsize 85.7}) & 57.3({\scriptsize 82.3}) & 43.7({\scriptsize 78.3}) & 54.3({\scriptsize 85.7}) & 51.6~{\tiny \color{green!60!black}(+7.7)} \\
        RL4VLA \citep{liu2025rlbringvlageneralization} & 93.0 ({\scriptsize 98.3}) & 91.3 ({\scriptsize 96.7}) & 92.0 ({\scriptsize 97.0}) & 93.7 ({\scriptsize 98.3}) & 92.5 \\
        \rowcolor{gray!10}
        SFT $\rightarrow$ \nameppo (\textcolor{tumblue}{\textbf{\namemodule}}) & 94.3 ({\scriptsize 97.7}) & 95.3 ({\scriptsize 99.0}) & 93.7  ({\scriptsize 98.3}) & 95.0 ({\scriptsize 98.7}) & 94.6~{\tiny \color{green!60!black}(+2.1)} \\
        \rowcolor{gray!15} 
        \textbf{IPI (\textcolor{tumblue}{\textbf{\namemodule}})} & \textbf{98.0} ({\scriptsize 99.0}) & \textbf{98.5} ({\scriptsize 99.5}) & \textbf{98.0} ({\scriptsize 99.0}) & \textbf{97.5} ({\scriptsize 99.0}) & \textbf{98.0} \\
        \midrule
        \multicolumn{6}{l}{\textit{Pi0.5 Based Methods}} \\
        SFT & 49.3({\scriptsize 85.3}) & 64.7({\scriptsize 89.3}) & 44.7({\scriptsize 76.0}) & 69.7({\scriptsize 92.3}) & 57.1 \\
        GRAPE \citep{zhang2024grape} & 48.0({\scriptsize 78.7}) & 59.3({\scriptsize 88.3}) & 48.3({\scriptsize 69.7}) & 58.7({\scriptsize 80.3}) & 53.6 \\
        \rowcolor{gray!5} 
        SFT $\rightarrow$ \nametpo (\textcolor{tumblue}{\textbf{\namemodule}}) & 54.0({\scriptsize 83.7}) & 65.3({\scriptsize 80.3}) & 54.0({\scriptsize 70.7}) & 68.7({\scriptsize 83.7}) & 60.5~{\tiny \color{green!60!black}(+6.9)} \\
    $\pi_{\texttt{RL}}$~\citep{chen2025pi_} & 82.7 ({\scriptsize 98.3}) & 97.3 ({\scriptsize 99.3}) & 83.3 ({\scriptsize 97.3}) & 55.0 ({\scriptsize 69.7}) & 79.6 \\
        \rowcolor{gray!10}
        SFT $\rightarrow$ \nameppo (\textcolor{tumblue}{\textbf{\namemodule}}) & 90.7 ({\scriptsize 98.7}) & 97.7 ({\scriptsize 99.0}) & 85.7  ({\scriptsize 97.3}) & 63.7 ({\scriptsize 85.7}) & 84.5~{\tiny \color{green!60!black}(+4.9)} \\
        \rowcolor{gray!15} 
        \textbf{IPI (\textcolor{tumblue}{\textbf{\namemodule}})} & \textbf{95.7} ({\scriptsize 99.3}) & \textbf{98.7} ({\scriptsize 99.7}) & \textbf{93.0} ({\scriptsize 98.0}) & \textbf{78.7} ({\scriptsize 91.0}) & \textbf{91.5} \\
        \bottomrule
    \end{tabularx}
    }
    \vspace{-3mm}
\end{table*}
\paragraph{Main Results.}
We begin by presenting overall comparisons on two widely used families of manipulation benchmarks. Table~\ref{tab:simplerenv_windowx} reports grasp and final success rates on the four representative \textbf{SimplerEnv-WidowX} tasks, while Table~\ref{tab:maniskill3} reports results on selected \textbf{ManiSkill3-Franka} tasks including both stacking and contact-rich manipulation. 

Across all benchmarks, existing VLA baselines exhibit limited performance 
(e.g., average success rates $<60\%$). Recent RL fine-tuning approaches, such as 
{RL4VLA}~\citep{liu2025rlbringvlageneralization} achieve strong results 
($92.5\%$ on SimplerEnv-WidowX, $70.5\%$ on ManiSkill3). 
Our proposed \textbf{IPI} further improves to $98.0\%$ and $96.4\%$, 
outperforming prior state-of-the-art methods by $+5.4$ and $+25.9$ points, nearly solving these benchmark tasks. 
Flow-based VLA models such as $\pi_{0.5}$ also benefit substantially from our stage-aware reinforcement design: integrating \namemodule{} consistently boosts $\pi_{0.5}$’s performance across all tasks, outperforming its original flow-matching baseline by a large margin.
Our \textbf{IPI} is a fully implemented and executed pipeline obtained from actual end-to-end runs,  demonstrating that each stage can be integrated seamlessly and that the complete framework achieves the strongest overall performance. 

\begin{table}[t]
    \centering
    \caption{\textbf{Evaluation on selected ManiSkill3 Franka tasks.} OpenVLA-7B and Pi0.5 based methods use 100 trajectory samples for SFT and 50 trajectory preference pairs for TPO and \nametpo{}, detailed in Appendix \ref{app:datacollection}.}
    \label{tab:maniskill3}
    \resizebox{0.93\textwidth}{!}{
    \begin{tabularx}{\linewidth}{>{\raggedright\arraybackslash}X ccccc}
        \toprule
        \multirow{2}{*}{\textbf{Methods}} & \multicolumn{4}{c}{\textbf{Robotic Task}} & \multirow{2}{*}{\makecell{\textbf{Avg.} \\ \textbf{Success} \\ \textbf{Rate (\%)} }} \\
        \cmidrule(lr){2-5}
        & \makecell{Stack \\ Cube} & \makecell{Push \\ Cube} & \makecell{Pull \\ Cube} & \makecell{LiftPeg \\ Upright} & \\
        
        \midrule
        \multicolumn{6}{l}{\textit{Other Methods}} \\
        Octo (fine-tuning)~\citep{ghosh2024octo} & 1.0 & 67.0 & 90.0 & 0.0 & 39.5 \\
        SmolVLA (fine-tuning)~\citep{shukor2025smolvlavisionlanguageactionmodelaffordable} & 12.7 & 86.3 & 90.7 & 16.3 & 51.5 \\
        RoboFAC-7B~\citep{lu2025robofaccomprehensiveframeworkrobotic} & 85.5 & 80.4 & 80.7 & 84.0 & 82.7 \\
        \midrule
        \multicolumn{6}{l}{\textit{OpenVLA-7B Based Methods}} \\
        SFT & 12.0 & 11.7 & 31.0 & 5.3 & 15.0 \\
        GRAPE~\citep{zhang2024grape} & 15.7 & 13.3 & 35.3 & 7.7 & 18.0 \\
        \rowcolor{gray!5} 
        SFT$\rightarrow$\nametpo{} (\textcolor{tumblue}{\textbf{\namemodule}}) & 19.3 & 16.0 & 35.7 & 12.3 & 20.8~{\tiny \color{green!60!black}(+2.8)} \\
        RL4VLA ~\citep{liu2025rlbringvlageneralization} & 64.0 & 95.7 & 90.3 & 32.0  & 70.5 \\
        \rowcolor{gray!10} 
        SFT$\rightarrow$\nameppo{} (\textcolor{tumblue}{\textbf{\namemodule}}) & 92.7 & 96.0 & 95.3 & 89.7 & 93.4~{\tiny \color{green!60!black}(+22.9)}\\
        \rowcolor{gray!15} 
        \textbf{IPI (\textcolor{tumblue}{\textbf{\namemodule}})} & \textbf{94.3} & \textbf{97.3} & \textbf{98.5} & \textbf{95.5} & \textbf{96.4}\\
        \midrule
        \multicolumn{6}{l}{\textit{Pi0.5 Based Methods}} \\
        SFT & 26.3 & 18.3 & 43.0 & 10.7 & 25.6 \\
        \rowcolor{gray!5} 
        GRAPE~\citep{zhang2024grape} & 22.7 & 16.3 & 45.0 & 6.3 & 22.6 \\
        SFT$\rightarrow$\nametpo{} (\textcolor{tumblue}{\textbf{\namemodule}}) & 28.0 & 22.3 & 44.7 & 15.7 & 27.7~{\tiny \color{green!60!black}(+5.1)} \\
    $\pi_{\texttt{RL}}$~\citep{chen2025pi_} & 72.3 & 96.7 & 93.3 & 58.0  & 80.1 \\
        \rowcolor{gray!10} 
        SFT$\rightarrow$\nameppo{} (\textcolor{tumblue}{\textbf{\namemodule}}) & 80.7 & 98.0 & 92.7 & 75.3 & 86.7~{\tiny \color{green!60!black}(+6.6)}\\
        \rowcolor{gray!15} 
        \textbf{IPI (\textcolor{tumblue}{\textbf{\namemodule}})} & \textbf{84.3} & \textbf{99.3} & \textbf{95.0} & \textbf{80.7} & \textbf{89.9}\\
        \bottomrule
    \end{tabularx}
    }
\end{table}
\begin{figure}[h]
  \centering
  \vspace{10pt}
  \begin{overpic}[width=0.95\textwidth,trim=0cm 0 0 0cm,clip]{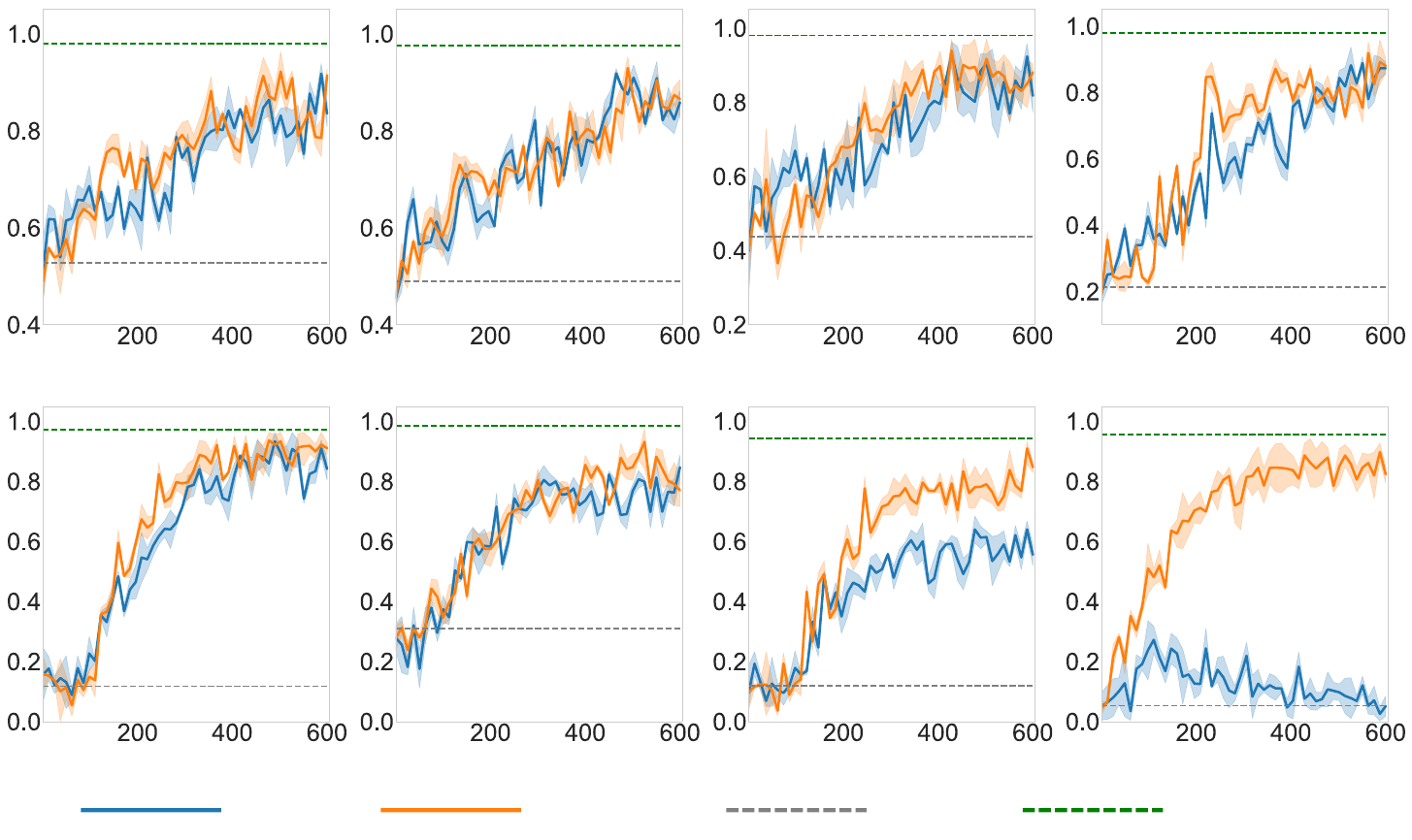}

    \put(-2.5,42.5){\rotatebox{90}{\footnotesize  \textbf{SimplerEnv}}}
    \put(-2.5,14.5){\rotatebox{90}{\footnotesize  \textbf{ManiSkill3}}}

    \put(5,60.5){\color{black}{\footnotesize  PutCarrotOnPlate}}
    \put(28.5,60.5){\color{black}{\footnotesize  PutEggplantInBasket}}
    \put(55,60.5){\color{black}{\footnotesize  PutSpoonOnTowel}}
    \put(77.5,60.5){\color{black}{\footnotesize  StackGreenOnYellow}}
    
    \put(8.3,32){\color{black}{\footnotesize  PushCube}}
    \put(33.8,32){\color{black}{\footnotesize  PullCube}}
    \put(58,32){\color{black}{\footnotesize  StackCube}}
    \put(80.5,32){\color{black}{\footnotesize  LiftPegUpright}}

    \put(17,1.5){\color{black}{\footnotesize  PPO}}
    \put(38,1.5){\color{black}{\footnotesize \nameppo{}}}
    \put(63,1.5){\color{black}{\footnotesize SFT}}
    \put(84,1.5){\color{black}{\footnotesize \textbf{IPI}}}
    
  \end{overpic}
  \vspace{-4px}
\caption{\small 
Comparison of learning curves across eight representative tasks from SimplerEnv and ManiSkill3. 
The y-axis denotes the success rate, and the x-axis shows the interaction environment steps (in thousands). 
}
\label{fig:8plots}
\vspace{-5mm}
\end{figure}

We compare SFT, PPO, \nameppo{}, and our full IPI method. While PPO improves over SFT, it often stagnates in high-precision (e.g., \emph{StackCube)} or contact-rich settings (e.g., \emph{LiftPegUpright}). 
\nameppo{} accelerates convergence and achieves higher asymptotic performance by leveraging stage-aware signals. 
Notably, the most challenging tasks, \emph{LiftPegUpright} and \emph{StackCube}, show the clearest benefit, highlighting the importance of incorporating stage-awareness for solving complex tasks.
\paragraph{PPO vs \nameppo{}}
While PPO improves over SFT, it often stagnates on tasks requiring high precision or contact-rich interactions. In contrast, \nameppo{} consistently accelerates convergence and achieves higher asymptotic performance by leveraging stage-aware signals. Figure~\ref{fig:8plots} presents results across eight representative tasks from SimplerEnv-WidowX and ManiSkill3. The most challenging tasks—\emph{LiftPegUpright} and \emph{StackGreenOnYellow}—exhibit the largest performance gaps, underscoring the importance of incorporating stage-aware signals in long-horizon, precision-critical manipulation. By comparison, for short-horizon pick-and-place tasks (e.g., \emph{PutCarrotOnPlate}, \emph{PutEggplantInBasket}) or simple push-and-pull tasks, PPO and \nameppo{} achieve similar final success rates, with \nameppo{} mainly contributing faster convergence and reduced variance. Overall, these results suggest that stage-aware guidance is particularly crucial when strict alignment accuracy or multi-stage coordination is required, whereas simpler tasks can often be solved effectively with standard reinforcement learning.

After the benchmark-level comparison in Tables~\ref{tab:simplerenv_windowx} 
and \ref{tab:maniskill3}, we note that while the overall improvements of \nameppo{} and \nametpo{} over prior baselines are consistent, the performance gap is most pronounced on two tasks: 
(1) Cube stacking tasks from both the environments, which requires precise alignment in placing stage, and (2) \emph{LiftPegUpright} from ManiSkill3, which demands accurate orientation control after lifting. 
To better understand where these gains originate, we decompose trajectories into semantic stages and evaluate \textbf{conditional stage success} 
(\(P(\text{stage}_k \mid \text{stage}_{k-1})\)), which measures how reliably a policy completes a stage given that all previous stages have been successful.

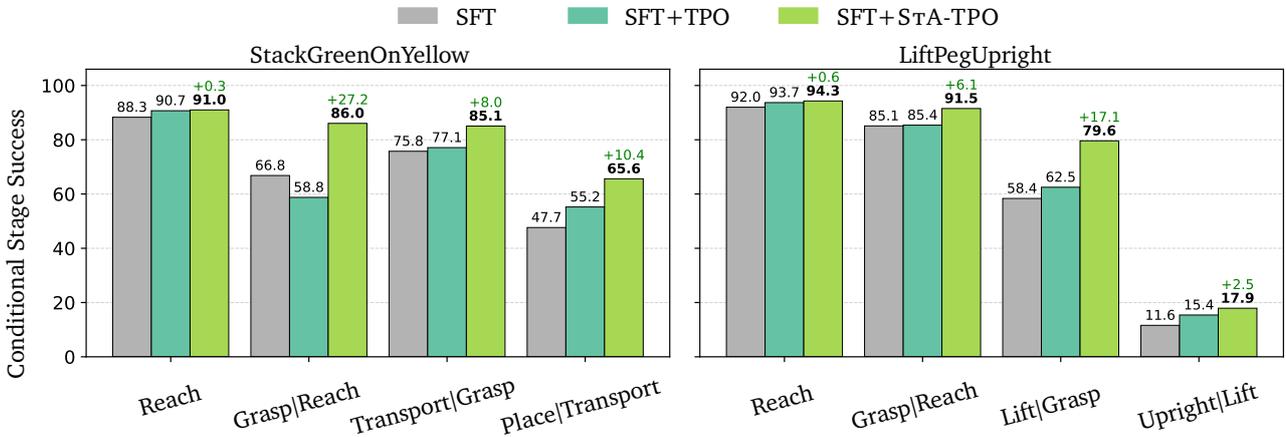
\begin{figure}[!h]
  \centering
  \vspace{5pt}
  \begin{overpic}[width=0.99\textwidth,trim=0cm 0 0 0cm,clip]{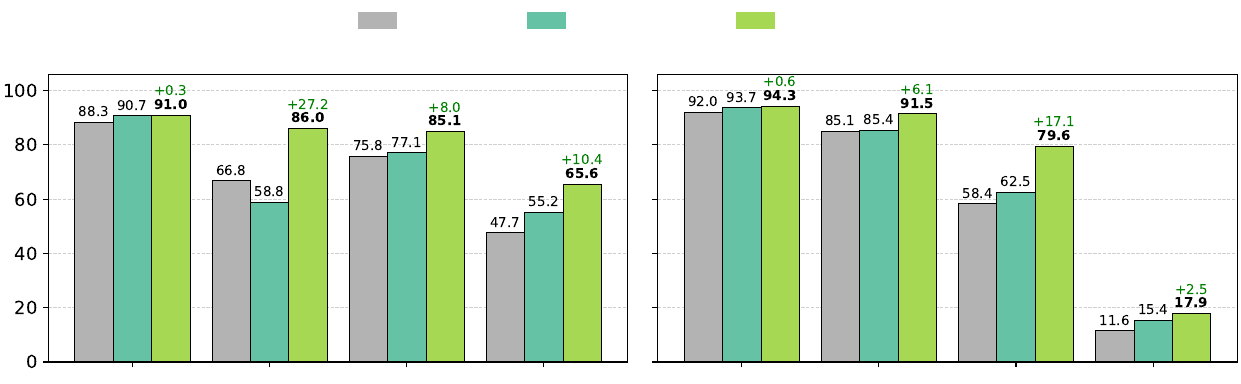}

    \put(-2.5,0){\rotatebox{90}{\footnotesize  Conditional Stage Success}}

    \put(33.5,28.5){{\footnotesize  SFT}}
    \put(47,28.5){{\footnotesize  SFT+TPO}}
    \put(64,28.5){{\footnotesize  SFT+\nametpo{}}}

    \put(17,25.5){{\footnotesize  {StackGreenOnYellow}}}
    \put(69,25.5){{\footnotesize  {LiftPegUpright}}}

    \put(8,-3){\rotatebox{15}{\footnotesize  Reach}}
    \put(15.5,-4.1){\rotatebox{15}{\footnotesize Grasp$|$Reach}}
    \put(25,-4.5){\rotatebox{15}{\footnotesize Transport$|$Grasp}}
    \put(37,-4.5){\rotatebox{15}{\footnotesize {Place$|$Transport}}}
    \put(57,-3){\rotatebox{15}{\footnotesize  Reach}}
    \put(65,-4.1){\rotatebox{15}{\footnotesize Grasp$|$Reach}}
    \put(77,-3.8){\rotatebox{15}{\footnotesize Lift$|$Grasp}}
    \put(88,-4.1){\rotatebox{15}{\footnotesize {Upright$|$Lift}}}

  \end{overpic}
  \vspace{20px} %15 to 20
    \caption{\small \textbf{Offline Stage-wise ablation on two tasks.} 
    We report stage completion rates (\%) for \emph{StackGreenonYellow} (SimplerEnv) 
    and \emph{LiftPegUpright} (ManiSkill3). 
    Compared with TPO, \nametpo{} 
    achieves significant gains, particularly in the \textbf{grasp} and 
    \textbf{place/upright} stages, which are critical for final success.
}
    \label{fig:stage_completion}
    \vspace{0mm}
\end{figure}

Figure~\ref{fig:stage_completion} shows that \nametpo{} provides clear advantages over TPO, with the largest improvements appearing in the \textbf{grasp},  and \textbf{place} and \textbf{upright} stages. These stages are particularly decisive for the final outcome, explaining why the overall success rate improvements are disproportionately large for these two tasks.
% \begin{figure}[t]
%     \centering
%     \includegraphics[width=0.95\linewidth]{images/stage_completion_two_tasks_set2_conditional.pdf}
%     \caption{\textbf{Offline Stage-wise ablation on two tasks.} 
%     We report stage completion rates for \emph{StackGreenonYellow} (SimplerEnv) 
%     and \emph{LiftPegUpright} (ManiSkill3). 
%     Compared with TPO, \nametpo{} 
%     achieves significant gains, particularly in the \textbf{grasp} and 
%     \textbf{place/upright} stages, which are critical for final success.}
%     \label{fig:stage_completion}
% \end{figure}

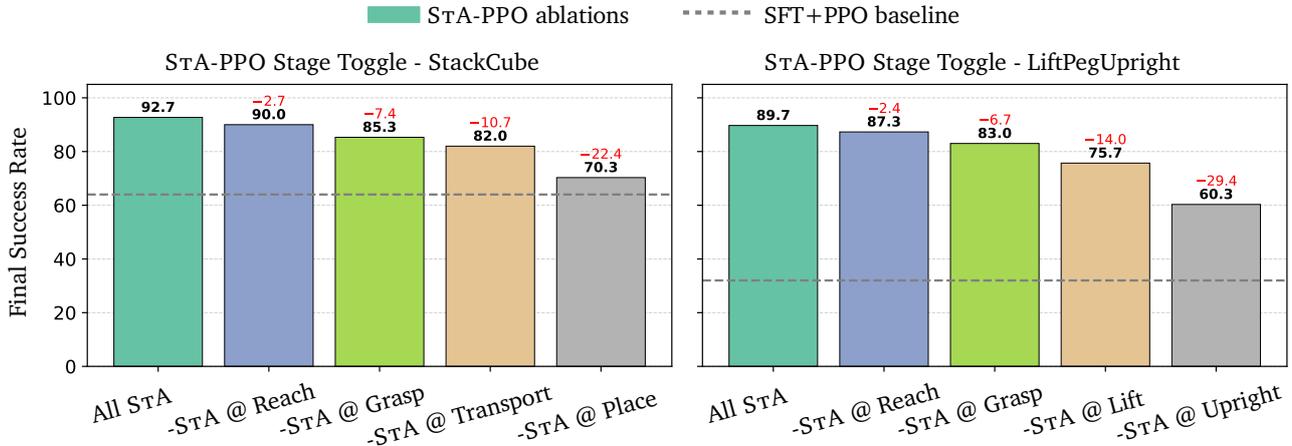
\begin{figure}[!h]
  \centering
  \vspace{3pt}
  \begin{overpic}[width=0.99\textwidth,trim=0cm 0 0 0cm,clip]{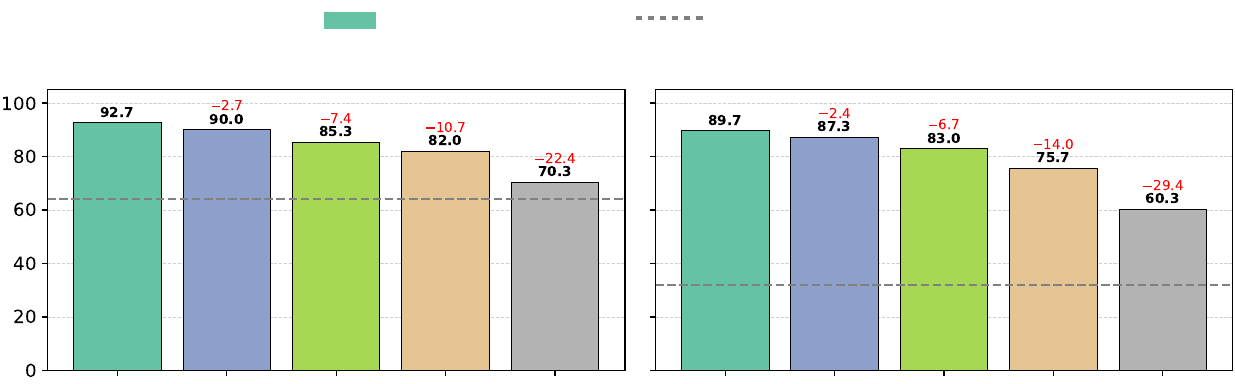}

    \put(-2.5,5){\rotatebox{90}{\footnotesize  Final Success Rate}}

    \put(31,28.8){{\footnotesize  \nameppo{} ablations}}
    \put(58,28.8){{\footnotesize  SFT+PPO baseline}}

    \put(10,25){{\footnotesize  \nameppo{} Stage Toggle - StackCube}}
    \put(58,25){{\footnotesize  \nameppo{} Stage Toggle - LiftPegUpright}}

    \put(4,-3.5){\rotatebox{15}{\footnotesize  All \textsc{StA}}}
    \put(10,-4.6){\rotatebox{15}{\footnotesize -\textsc{StA} @ Reach}}
    \put(19,-4.6){\rotatebox{15}{\footnotesize -\textsc{StA} @ Grasp}}
    \put(26,-5.5){\rotatebox{15}{\footnotesize {-\textsc{StA} @ Transport}}}
    \put(38,-5){\rotatebox{15}{\footnotesize  -\textsc{StA} @ Place}}

    \put(53.3,-3.5){\rotatebox{15}{\footnotesize  All \textsc{StA}}}
    \put(60,-4.6){\rotatebox{15}{\footnotesize -\textsc{StA} @ Reach}}
    \put(68.5,-4.6){\rotatebox{15}{\footnotesize -\textsc{StA} @ Grasp}}
    \put(78.5,-4.6){\rotatebox{15}{\footnotesize {-\textsc{StA} @ Lift}}}
    \put(86,-5){\rotatebox{15}{\footnotesize  -\textsc{StA} @ Upright}}

  \end{overpic}
  \vspace{20px} %15 to 20
    \caption{\small \textbf{Stage toggle ablation of \nameppo{}.} 
    We evaluate the effect of selectively disabling stage-aware reinforcement signals 
    on two representative tasks: \emph{StackCube} (ManiSkill3) 
    and \emph{LiftPegUpright} (ManiSkill3). 
    The \textbf{All \textsc{\textsc{StA} (\namemodule)}} setting achieves the best performance, while disabling critical stages (\textbf{Place} in stacking, 
    \textbf{Upright} in peg lifting) causes the largest performance drops. 
}
    \label{fig:stappo_toggle}
    \vspace{-3mm}
\end{figure}

\paragraph{Ablations study}
To further dissect the contributions of stage-aware reinforcement, we conduct a stage toggle ablation where the \namemodule{} signal is selectively removed at different phases of the manipulation in \nameppo{}. As shown in Figure~\ref{fig:stappo_toggle}, disabling \namemodule{} at early stages (e.g., reach or grasp) only leads to moderate drops, since later corrective actions can partially recover performance. In contrast, removing \namemodule{}{} at the final precision-critical phases (e.g., \textbf{Place} in stacking and \textbf{Upright} in peg lifting) causes the largest degradation, reducing success rates by more than 20\%. This analysis highlights that \namemodule{} guidance is especially valuable at stages where geometric accuracy and stability directly determine task completion.

% \begin{figure}[t]
%     \centering
%     \includegraphics[width=0.95\linewidth]{images/sta_ppo_stage_toggle_with_pctdrop.pdf}
%     \caption{\textbf{Stage toggle ablation of \nameppo{}.} 
%     We evaluate the effect of selectively disabling stage-aware reinforcement signals 
%     on two representative tasks: \emph{Stack Green on Yellow} (SimplerEnv) 
%     and \emph{LiftPeg Upright} (ManiSkill3). 
%     The \textbf{All STA} setting achieves the best performance, while 
%     disabling critical stages (\textbf{Place} in stacking, 
%     \textbf{Upright} in peg lifting) causes the largest performance drops. 
%     A gray dashed line denotes the \textbf{SFT+PPO baseline} for reference.}
%     \label{fig:stappo_toggle}
% \end{figure}

% We begin with a teaser comparison to highlight the performance gap between existing VLA baselines and our RL-based method.
% Table~\ref{tab:simplerenv_windowx} reports grasp and final success rates on four representative \textbf{SimplerEnv-WidowX} tasks.
% Classical VLA baselines exhibit limited success, with averages ranging from $1.1\%$ to $58.3\%$.
% Recent RL fine-tuning approaches, such as \textbf{RL4VLA}~\citep{liu2025rlbringvlageneralization}, achieve much stronger results (average $92.6\%$).
% However, our proposed \textbf{IPI} further improves to an average success rate of $98.0\%$, a $+5.4$ point gain over RL4VLA, nearly solving the benchmark.
% \input{tables/simplerenv_widowx}

% \input{tables/simpler_ood}

% \paragraph{Ablations.}
% We study (i) object-aware vs.\ spatial-aware STPO pairs, (ii) temperature $\beta$ in Eq.~(\ref{eq:stpo}), and (iii) initializing PPO's critic from the STPO-trained value head. 
% Ablations indicate that (a) combining both STPO pair types yields the strongest OOD gains, and (b) value/advantage initialization improves PPO stability (Appendix).

% \newpage

\section{Conclusion}
We presented \emph{Stage-Aware Reinforcement} (\namemodule{}), a plug-in module that decomposes trajectories into semantically meaningful stages and provides stage-level reinforcement signals. Building on this, we introduced Stage-Aware TPO (\nametpo{}) and PPO (\nameppo{}) for offline stage-wise preference alignment and online intra-stage interaction, and integrated them with supervised fine-tuning into the \emph{Imitation $\to$ Preference $\to$ Interaction} (IPI) pipeline. Experiments on SimplerEnv and ManiSkill3 demonstrate that IPI achieves state-of-the-art success rates.

Extensive experiments on both OpenVLA and Pi0.5 backbones, across SimplerEnv and ManiSkill3, show that IPI consistently yields substantial gains and achieves state-of-the-art success rates. These results demonstrate that stage-aware credit assignment is a key missing ingredient in current VLA reinforcement learning, enabling more stable optimization, faster convergence, and markedly stronger long-horizon performance. We believe this stage-centric perspective opens up promising directions for scalable, interpretable, and robust robotic learning.

\newpage

\bibliography{main}

\newpage

\appendix
\appendix

\clearpage

\section{Training Settings and Evaluation Protocols}
\label{app:training}

\subsection{Supervised Fine-Tuning (SFT)}
We initialize SFT models from two pretrained VLA backbones: OpenVLA-7B and the pi0.5\_base. For each backbone, we optimize the action prediction objective using the AdamW optimizer with a learning rate of $1\times 10^{-5}$ and a constant schedule. SFT is trained for 100K steps per backbone on each benchmark, and the same training budget is used for all methods on the same backbone to ensure fair comparison. All image observations are resized to $224 \times 224$ before being fed into the VLA backbone.

\subsection{Trajectory Preference Optimization: TPO and STA-TPO}
For offline preference optimization, we start from SFT checkpoints of both OpenVLA-7B and pi0.5\_base backbone, and construct datasets of paired successful (``chosen'') and failed (``rejected'') trajectories. Each trajectory consists of image observations, language instructions, and continuous robot actions, which are discretized into tokens using an action tokenizer built on top of the base vocabulary.

We use a DPO-style preference objective that increases the log-likelihood ratio between chosen and rejected trajectories relative to a frozen SFT reference model. This provides a stable and
scalable way to perform trajectory-level preference alignment. Training uses AdamW with a learning rate of $2\times10^{-5}$, gradient accumulation, and 50K optimization steps per backbone to ensure fair comparison across baselines.

\nametpo~uses the same data pipeline, objective, and optimizer, but modifies the DPO logit difference by adding stage-aware margins computed from our stage calculator. This biases the
preference updates toward stages where the chosen and rejected trajectories differ the most, while keeping easy stages largely unchanged, yielding a stage-aware variant that is directly
comparable to standard TPO.

\subsection{Reinforcement Learning: PPO and STA-PPO}
We adopt an on-policy PPO framework. For SimplerEnv, we run 100 parallel environments with an episode horizon of 60 steps, yielding
6000 transitions per PPO update. Rollouts are stored in a separated replay buffer, and advantages are computed using generalized advantage estimation with a discount factor $\gamma = 0.99$ and GAE parameter $\lambda = 0.95$.

Optimization follows the standard clipped PPO objective. We use the AdamW optimizer with a policy learning rate of $1\times10^{-4}$, a value learning rate of $3\times10^{-3}$, and gradient accumulation over 20 steps to stabilize large-model training. Each update performs one PPO epoch over four minibatches, and the entropy coefficient is set to 0.0. Unless otherwise stated, ManiSkill3 uses the same optimization settings and training budget. All runs are trained for
600K environment steps to ensure fair comparison across baselines.

\nameppo~follows the same training pipeline and computational budget but augments the reward with dense stage-aware signals generated by the stage calculator. These structured shaping terms provide intermediate guidance that complements sparse task rewards, and the system additionally logs per-stage performance statistics during rollouts. Aside from this reward augmentation, \nameppo~is identical to PPO, enabling controlled comparisons of stage-aware credit assignment.

\subsection{Evaluation Protocols}
We evaluate all methods on both SimplerEnv-WidowX and ManiSkill3 benchmarks under a unified protocol.  
Each policy is tested over 300 evaluation episodes with deterministic decoding. A rollout is considered successful if the environment-defined task condition is satisfied within the evaluation horizon (60 steps for SimplerEnv and 30 steps for ManiSkill3).  
Results are averaged over three random seeds. In addition to final success rates, we compute conditional stage success to diagnose at which manipulation stages policies succeed or fail.

\subsection{Checkpoint Selection}
We adopt a fixed-duration training schedule for all methods and use a consistent checkpoint-selection protocol across backbones and benchmarks.

For TPO and \nametpo, models are trained for 50K optimization steps, and evaluated every 1{,}000 training steps on both SimplerEnv and ManiSkill3. The best-performing checkpoint under this evaluation protocol is reported.

For PPO and \nameppo, models are trained for 600K steps. We run periodic evaluations every 6{,}000 environment steps on SimplerEnv and every 3{,}000 environment steps on ManiSkill3, selecting the checkpoint with the highest success rate.

For all ablation studies, we fix the total training duration and report the final checkpoint to ensure strict comparability across variants.

\section{Experimental Setup and Implementation Details}
\label{sec:bench_details}

\subsection{Environments and Task Definitions}

To evaluate robotic policies under diverse manipulation scenarios, we conduct experiments in two simulation environments: \textbf{SimplerEnv-WidowX} and \textbf{ManiSkill3-Franka}.

\subsubsection{SimplerEnv-WidowX}  
We adopt the benchmark provided by SimplerEnv~\citep{li2024evaluating}. Within this environment, we test the following tasks:

\begin{itemize}
  \item \textbf{Put the spoon on the towel} — The spoon is initialized at one corner of a \(15\!\times\!15\text{\,cm}^2\) square region on the tabletop; a towel is placed at the opposite corner. The spoon’s orientation alternates between horizontal and vertical across trials, requiring the gripper to adapt its pose accordingly.  
  \item \textbf{Put the carrot on the plate} — This task mirrors the previous one, with a carrot replacing the spoon and a plate replacing the towel.  
  \item \textbf{Stack the green block on the yellow block} — A green block and a yellow block (both \(3\;\text{cm}\) cubes) are placed at different corners of a tabletop square. We evaluate two square sizes (side length 10 cm and 20 cm). This task tests precise grasping, lifting, and alignment to form a stable stack.  
  \item \textbf{Put the eggplant into the yellow basket} — An eggplant is randomly placed in the right basin of a simulated sink, and a yellow basket is placed in the left basin. The eggplant’s position and orientation vary across trials (while avoiding basin edges to ensure graspability). This task challenges the policy’s ability to deal with irregular shapes and variable object poses.
\end{itemize}

\subsubsection{ManiSkill3-Franka}  
To further stress-test manipulation under contact-rich and long-horizon tasks, we use four tasks from ManiSkill3~\citep{taomaniskill3} with a simulated Franka robot:

\begin{itemize}
  \item \textbf{Stack Cube} — The robot must stack one cube on top of another, requiring accurate grasping, lifting, alignment, and stable placement.  
  \item \textbf{Push Cube} — A cube is placed on the table, and the robot must push it toward a designated target region. This task emphasizes smooth trajectory execution and contact-rich control.  
  \item \textbf{Pull Cube} — The robot grasps the cube on one side and pulls it (dragging) into the target region, demanding stable grasp maintenance under sliding contact.  
  \item \textbf{Lift Peg Upright} — A peg is initialized lying flat on the table; the robot must grasp it, lift it, and reorient it so that it stands vertically upright. This task is challenging due to orientation constraints and the need for precise control to maintain balance.
\end{itemize}

\subsection{Evaluation Protocol}

For all experiments, we follow the evaluation procedure below:

\begin{itemize}
  \item Each task is repeated across the full set of defined trials. For SimplerEnv we run 300 trials per task; for ManiSkill3 we follow the standard configurations defined by the benchmark.  
  \item Success is determined based on task-specific completion criteria (e.g., correct placement, stable stack, upright peg, etc.).  
  \item For tasks (or sequences) that involve multiple steps, we additionally record the trajectory length and measure efficiency, robustness, and consistency across trials.  
  \item Final results (success rates, trajectory lengths) are reported as mean ± standard deviation over all trials, ensuring statistical reliability.
\end{itemize}

\subsection{Implementation Details}

\begin{itemize}
  \item Both SimplerEnv and ManiSkill3 are used with their default simulation configurations. We do not modify the underlying environment logic — only vary object initial positions and orientations as described.  
  \item For each trial, object poses (positions, orientations when applicable) are randomized within the constraints defined for each task to ensure diversity across trials.  
  \item Policies operate in an end-to-end fashion (observations → actions), without manual heuristics or task-specific engineering.  
  \item All reported metrics are averaged over the full set of trials. When appropriate, we also report standard deviations to reflect variability.  
\end{itemize}

% If you have additional training details (e.g. hyper-parameters, network architecture, input modalities), you can add another subsection here.

\subsection{Data Collection}
\label{app:datacollection}
Our data collection process, which is similar to the methodology used in the main paper, is designed to generate high-quality data for both supervised and preference-based learning. All data is collected specifically for the selected ManiSkill3 tasks.

\begin{itemize}
    \item \textbf{Expert Demonstrations:} For each task, we generate \textbf{100 high-quality demonstration trajectories}. These trajectories are produced using the MPLib motion planner, ensuring kinematically feasible and efficient paths to task completion. Following the findings of the main paper, we apply an action filtering technique to this data, removing idle actions where the end-effector pose changes by a negligible amount. This preprocessing step is crucial for mitigating the issue of trained SFT policies getting stuck during execution.

     \item \textbf{Preference Pairs:} 
        For methods requiring preference data (e.g., TPO), we generate \textbf{50 trajectory preference pairs} per task. In the case of SimplerEnv, trajectories are sampled for each task using the Octo model.
        These pairs are obtained by sampling two trajectories from the Octo-collected dataset, and assigning preference labels based on cumulative rewards (e.g., successful completion vs. failure). 
\end{itemize}

% \begin{figure}[H]
% \centering
% \includegraphics[width=\textwidth]{images/trajectory_plot_no_line.pdf}
% \caption{\textbf{Four-stage decomposition across manipulation tasks.} Each task naturally decomposes into four semantic stages, with stage durations reflecting actual task dynamics.}
% \label{fig:stages}
% \end{figure}

% \input{tables/star-config}

\section{Stage-wise Cost and Potential Definitions}
\label{app:stagewisecostandpotential}

This appendix provides the complete definitions of the stage-wise costs
$\ell_k(\tau^{(k)})$ and progress potentials $\Phi_k(s)$ used in \namemodule.
Each stage corresponds to a geometric manipulation primitive.
Stage costs---used in \nametpo---evaluate how well a trajectory segment
achieves the geometric goal of a stage, while stage potentials---used in
\nameppo---provide dense, per-step shaping signals that reflect normalized
progress within a stage. All potentials follow a normalized sigmoid form,
\[
\sigma(z)=\frac{1}{1+e^{-z}},
\]
and lie in $[0,1]$. For stages involving orientation, we additionally use a
normalized rotational alignment measure $g(R_1,R_2)\in[0,1]$, where $g(R_1,R_2)$ denotes a normalized orientation alignment measure.

\paragraph{Stage Decomposition Across Tasks.}
Different manipulation tasks instantiate different subsets of these generic manipulation primitives. We use a minimal, rule-based decomposition based on geometric event boundaries that can be reliably detected from state
observations:

\begin{itemize}[leftmargin=1.5em]
    \item \textbf{Pick--Place:}
    Reach $\rightarrow$ Grasp $\rightarrow$ Transport $\rightarrow$ Place.

    \item \textbf{Push:}
    Reach $\rightarrow$ Push $\rightarrow$ Goal.

    \item \textbf{Pull:}
    Reach $\rightarrow$ Pull $\rightarrow$ Goal.

    \item \textbf{Peg Upright:}
    Reach $\rightarrow$ Grasp $\rightarrow$ Lift $\rightarrow$ Upright.
\end{itemize}

Each stage is associated with a well-defined geometric objective (e.g., approach, alignment, elevation, planar displacement, fine-grained goal adjustment).
The following sections list the complete cost and potential functions for all stages used in \namemodule~.

% ----------------------------------------------
\subsection{Reach}

\paragraph{Goal.}
Guide the end-effector toward the target object until it enters the vicinity
where grasp alignment becomes feasible. This stage captures the coarse
approach motion before any contact or fine-grained adjustment occurs.

\paragraph{Cost.}
We quantify reaching quality using the Euclidean distance between the
end-effector position and the object’s grasp point:
\[
d_{\mathrm{reach}}(t)
= \left\| x^{\mathrm{ee}}(t) - x^{\mathrm{obj}}(t) \right\|.
\]
The stage cost is the average approach error over the stage segment:
\[
\ell_{\mathrm{Reach}}(\tau^{(k)}) 
= \frac{1}{T_k}\sum_{t\in\tau^{(k)}} d_{\mathrm{reach}}(t).
\]

\paragraph{Potential.}
To provide dense shaping during the approach, we define a normalized potential
that increases as the end-effector draws closer to the object:
\[
\Phi_{\mathrm{Reach}}(s)
= \sigma\!\left(
1 - \frac{d_{\mathrm{reach}}(s)}{d_{\mathrm{reach}}^{\max}}
\right).
\]
Here $d_{\mathrm{reach}}^{\max}$ is a geometry-derived normalization scale,
chosen as the object’s characteristic size:
\[
d_{\mathrm{reach}}^{\max} = L_{\mathrm{obj}},
\]
where $L_{\mathrm{obj}}$ denotes the object’s side length (e.g., cube size in ManiSkill3)
or, more generally, its bounding-box diameter.  
The same $d_{\mathrm{reach}}^{\max}$ is used as the threshold for the
Reach~$\rightarrow$~Grasp transition, ensuring that approach shaping and stage
segmentation operate on a consistent geometric scale.

% \paragraph{Notation.}
% \begin{itemize}[leftmargin=1.5em]
%     \item $x^{\mathrm{ee}}(t)$: end-effector position at time $t$.
%     \item $x^{\mathrm{obj}}(t)$: object grasp-point position at time $t$.
%     \item $d_{\mathrm{reach}}(t)$: TCP--object distance during the Reach stage.
%     \item $T_k$: number of steps in the $k$-th stage of the trajectory.
%     \item $L_{\mathrm{obj}}$: characteristic object size (e.g., side length of cubes or
%     bounding-box diameter for general objects).
%     \item $d_{\mathrm{reach}}^{\max}$: normalization scale for the Reach potential,
%     set to $L_{\mathrm{obj}}$.
% \end{itemize}

% ----------------------------------------------
\subsection{Grasp}

\paragraph{Goal.}
Establish and maintain a stable grasp. In ManiSkill environments, the end-effector approaches the object,
aligns with the grasp point, and closes the gripper to securely hold the object.

\paragraph{Cost.}
The Grasp stage evaluates how well the end-effector pose matches the intended grasp configuration.
We measure this using the distance between the tool-center point (TCP) and the object’s grasp point:
\[
d_{\mathrm{pose}}(t)
= \left\| x^{\mathrm{tcp}}(t) - x^{\mathrm{obj}}(t) \right\|.
\]
The stage cost is the average geometric misalignment:
\[
\ell_{\mathrm{Grasp}}(\tau^{(k)})
= \frac{1}{T_k}\sum_{t \in \tau^{(k)}} d_{\mathrm{pose}}(t).
\]

\paragraph{Potential.}
We define a normalized potential that increases as the TCP approaches the grasp point:
\[
\Phi_{\mathrm{Grasp}}(s)
= \sigma\!\left(
1 - \frac{d_{\mathrm{pose}}(s)}{d_{\mathrm{grasp}}}
\right),
\]
where $\sigma(\cdot)$ is the logistic sigmoid and $d_{\mathrm{grasp}}$ is a rule-based normalization scale determined by object geometry:
\[
d_{\mathrm{grasp}} = L_{\mathrm{obj}},
\]
with $L_{\mathrm{obj}}$ the characteristic object size (e.g., the cube side length in ManiSkill3).
This scale is also used as the threshold for the Reach~$\rightarrow$~Grasp stage transition,
ensuring consistent geometry-aware shaping without per-task tuning.

% \paragraph{Notation.}
% \begin{itemize}[leftmargin=1.5em]
%     \item $x^{\mathrm{tcp}}(t)$: position of the robot’s tool-center point at time $t$.
%     \item $x^{\mathrm{cube}}(t)$: grasp point (center position) of the manipulated object.
%     \item $d_{\mathrm{pose}}(t)$: TCP--object distance at time $t$.
%     \item $T_k$: number of steps in the $k$-th stage segment $\tau^{(k)}$.
%     \item $L_{\mathrm{obj}}$: object characteristic size (e.g., side length of the cube); used for
%     both stage segmentation and potential normalization.
%     \item $d_{\mathrm{grasp}}$: grasp normalization scale, set to $L_{\mathrm{obj}}$.
% \end{itemize}

% ----------------------------------------------
\subsection{Transport}

\paragraph{Goal.}
Move the grasped object toward its target goal pose while maintaining a stable grasp.
This stage captures the coarse relocation of the object once it has been lifted or secured.

\paragraph{Cost.}
Transport quality is measured using the distance between the object’s current position
and the goal location:
\[
d_{\mathrm{trans}}(t)
= \left\| x^{\mathrm{obj}}(t) - x^{\mathrm{goal}} \right\|.
\]
The stage cost averages this residual goal error:
\[
\ell_{\mathrm{Transport}}(\tau^{(k)})
= \frac{1}{T_k}
\sum_{t \in \tau^{(k)}} d_{\mathrm{trans}}(t).
\]

\paragraph{Potential.}
To provide dense shaping toward the goal, we define:
\[
\Phi_{\mathrm{Transport}}(s)
= \sigma\!\left(
1 -
\frac{d_{\mathrm{trans}}(s)}{d_{\mathrm{trans}}^{\max}}
\right).
\]
The normalization scale $d_{\mathrm{trans}}^{\max}$ is set to the characteristic
task displacement:
\[
d_{\mathrm{trans}}^{\max}
= \left\| x^{\mathrm{obj}}_{\mathrm{init}} - x^{\mathrm{goal}} \right\|,
\]
i.e., the distance between the object’s initial position and its goal.
This provides a rule-based scale reflecting the required transport distance, and aligns potential shaping with the physical extent of the manipulation.

% \paragraph{Notation.}
% \begin{itemize}[leftmargin=1.5em]
%     \item $x^{\mathrm{obj}}(t)$: object position at time $t$.
%     \item $x^{\mathrm{goal}}$: goal position for the object.
%     \item $d_{\mathrm{trans}}(t)$: residual distance to the goal.
%     \item $T_k$: number of steps in the $k$-th stage.
%     \item $x^{\mathrm{obj}}_{\mathrm{init}}$: object position at the start of the episode.
%     \item $d_{\mathrm{trans}}^{\max}$: normalization scale for shaping, defined as the initial
%     object–goal separation.
% \end{itemize}

% ----------------------------------------------
\subsection{Place}

\paragraph{Goal.}
Position the object precisely at the goal location and stabilize it. This stage 
captures the fine-grained alignment after coarse transport has moved the 
object near its target.

\paragraph{Cost.}
Placement quality is measured by the residual distance between the object's 
current position and the goal:
\[
d_{\mathrm{place}}(t)
= \left\| x^{\mathrm{obj}}(t) - x^{\mathrm{goal}} \right\|.
\]
The stage cost averages this near-goal deviation:
\[
\ell_{\mathrm{Place}}(\tau^{(k)})
= \frac{1}{T_k}
\sum_{t \in \tau^{(k)}} d_{\mathrm{place}}(t).
\]

\paragraph{Potential.}
We define a normalized potential encouraging precise placement:
\[
\Phi_{\mathrm{Place}}(s)
= \sigma\!\left(
1 -
\frac{d_{\mathrm{place}}(s)}{d_{\mathrm{place}}^{\max}}
\right).
\]
The normalization scale is chosen as the object’s characteristic size:
\[
d_{\mathrm{place}}^{\max} = L_{\mathrm{obj}},
\]
which reflects the resolution required for fine positioning and provides a rule-based geometric scale.

% \paragraph{Notation.}
% \begin{itemize}[leftmargin=1.5em]
%     \item $x^{\mathrm{obj}}(t)$: object position at time $t$.
%     \item $x^{\mathrm{goal}}$: desired goal position.
%     \item $d_{\mathrm{place}}(t)$: near-goal positional error.
%     \item $L_{\mathrm{obj}}$: characteristic object dimension (e.g., cube side length).
%     \item $d_{\mathrm{place}}^{\max}$: normalization scale, set to $L_{\mathrm{obj}}$.
%     \item $T_k$: number of steps in the $k$-th stage.
% \end{itemize}

% ----------------------------------------------
\subsection{Push \& Pull}

\paragraph{Goal.}
Move the object toward the target goal while maintaining stable contact with the
end-effector. This stage unifies push- and pull-based planar manipulation, as both
correspond to contact-induced object translation in the plane.

\paragraph{Cost.}
Progress is measured using the residual distance between the object and its target:
\[
d_{\mathrm{pp}}(t)
= \left\| x^{\mathrm{obj}}(t) - x^{\mathrm{goal}} \right\|.
\]
The stage cost averages this residual error:
\[
\ell_{\mathrm{Push\&Pull}}(\tau^{(k)})
=
\frac{1}{T_k}
\sum_{t \in \tau^{(k)}} d_{\mathrm{pp}}(t).
\]

\paragraph{Potential.}
We define a normalized potential that increases as the object approaches the goal:
\[
\Phi_{\mathrm{Push\&Pull}}(s)
=
\sigma\!\left(
1 -
\frac{d_{\mathrm{pp}}(s)}{d_{\mathrm{pp}}^{\max}}
\right),
\]
where the normalization scale is chosen as the required planar displacement:
\[
d_{\mathrm{pp}}^{\max}
=
\left\| x^{\mathrm{obj}}_{\mathrm{init}} - x^{\mathrm{goal}} \right\|.
\]
This yields a rule-based geometric scale for shaping and is consistent with the
Transport stage, differing only by the presence of contact.

% \paragraph{Notation.}
% \begin{itemize}[leftmargin=1.5em]
%     \item $x^{\mathrm{obj}}(t)$: object position at time $t$.
%     \item $x^{\mathrm{goal}}$: target goal position in the plane.
%     \item $d_{\mathrm{pp}}(t)$: object--goal residual distance.
%     \item $x^{\mathrm{obj}}_{\mathrm{init}}$: initial object position at episode start.
%     \item $d_{\mathrm{pp}}^{\max}$: normalization scale, defined as the initial object--goal distance.
%     \item $T_k$: number of steps in the $k$-th stage segment.
% \end{itemize}

% ----------------------------------------------
\subsection{Lift}

\paragraph{Goal.}
Lift the object from the table to a target height $z_{\mathrm{goal}}$, ensuring
that it is safely elevated above obstacles for subsequent manipulation.

\paragraph{Cost.}
We define the lift deviation as the vertical residual:
\[
d_{\mathrm{lift}}(t)
= \left| z_{\mathrm{obj}}(t) - z_{\mathrm{goal}} \right|.
\]
The stage cost averages this error:
\[
\ell_{\mathrm{Lift}}(\tau^{(k)})
=
\frac{1}{T_k}
\sum_{t \in \tau^{(k)}}
d_{\mathrm{lift}}(t).
\]

\paragraph{Potential.}
We define a normalized shaping potential:
\[
\Phi_{\mathrm{Lift}}(s)
=
\sigma\!\left(
1 - 
\frac{
|z_{\mathrm{obj}}(s) - z_{\mathrm{goal}}|
}{
d_{\mathrm{lift}}^{\max}
}
\right),
\]
where
\[
d_{\mathrm{lift}}^{\max}
=
z_{\mathrm{goal}} - z_{\mathrm{table}}
\]
is the required lifting displacement. This provides a rule-based geometric scale and yields dense shaping that increases smoothly as the object approaches its target height.

% \paragraph{Notation.}
% \begin{itemize}[leftmargin=1.5em]
%     \item $z_{\mathrm{obj}}(t)$: object height at time $t$.
%     \item $z_{\mathrm{goal}}$: target lifting height.
%     \item $z_{\mathrm{table}}$: table height (environment constant).
%     \item $d_{\mathrm{lift}}(t)$: height residual during Lift.
%     \item $d_{\mathrm{lift}}^{\max} = z_{\mathrm{goal}} - z_{\mathrm{table}}$: normalization scale.
%     \item $T_k$: number of steps in the $k$-th stage.
% \end{itemize}

% ----------------------------------------------
\subsection{Upright}

\paragraph{Goal.}
Align the object's orientation with an upright target orientation.
This stage captures fine-grained rotational adjustment after the object has been lifted
or placed near its desired pose.

\paragraph{Cost.}
We measure uprightness using the angular deviation between the object orientation
$R_{\mathrm{obj}}(t)$ and the target upright orientation $R_{\mathrm{upright}}$.
Let
\[
d_{\mathrm{upright}}(t)
= \arccos\!\left(
\frac{
\operatorname{tr}\!\left(R_{\mathrm{upright}}^\top R_{\mathrm{obj}}(t)\right)-1
}{2}
\right),
\]
which equals the geodesic rotation distance on $\mathrm{SO}(3)$.
The stage cost averages this orientation error:
\[
\ell_{\mathrm{Upright}}(\tau^{(k)})
= \frac{1}{T_k}
\sum_{t \in \tau^{(k)}}
d_{\mathrm{upright}}(t).
\]

\paragraph{Potential.}
We define a normalized shaping potential that increases as the object approaches
its upright orientation:
\[
\Phi_{\mathrm{Upright}}(s)
=
\sigma\!\left(
1 -
\frac{d_{\mathrm{upright}}(s)}{d_{\mathrm{upright}}^{\max}}
\right).
\]
The normalization scale $d_{\mathrm{upright}}^{\max}$ corresponds to the maximum
possible rotational deviation:
\[
d_{\mathrm{upright}}^{\max} = \pi,
\]
which is the geodesic diameter of $\mathrm{SO}(3)$.
This choice yields a geometry-grounded scale that applies to any
upright-orientation task.

% \paragraph{Notation.}
% \begin{itemize}[leftmargin=1.5em]
%     \item $R_{\mathrm{obj}}(t)$: object orientation at time $t$.
%     \item $R_{\mathrm{upright}}$: desired upright orientation.
%     \item $d_{\mathrm{upright}}(t)$: geodesic rotation distance on $\mathrm{SO}(3)$.
%     \item $d_{\mathrm{upright}}^{\max} = \pi$: maximal rotation deviation.
%     \item $T_k$: number of steps in the $k$-th stage.
% \end{itemize}

% ----------------------------------------------
\subsection{Goal (Push/Pull)}

\paragraph{Goal.}
Ensure that the object reaches the target goal region and remains stably within it.

\paragraph{Cost.}
We measure goal attainment using the residual distance between the object and
the goal:
\[
d_{\mathrm{goal}}(t)
= \left\| x^{\mathrm{obj}}(t) - x^{\mathrm{goal}} \right\|.
\]
The stage cost averages this error:
\[
\ell_{\mathrm{Goal}}(\tau^{(k)})
=
\frac{1}{T_k}
\sum_{t \in \tau^{(k)}}
d_{\mathrm{goal}}(t).
\]

\paragraph{Potential.}
A normalized potential provides dense shaping near the goal:
\[
\Phi_{\mathrm{Goal}}(s)
=
\sigma\!\left(
1 -
\frac{d_{\mathrm{goal}}(s)}{d_{\mathrm{goal}}^{\max}}
\right),
\]
where
\[
d_{\mathrm{goal}}^{\max} = L_{\mathrm{obj}},
\]
the characteristic object size. This creates a fine-scale potential landscape
around the goal region, allowing smooth convergence without relying on a binary
indicator.

% \paragraph{Notation.}
% \begin{itemize}[leftmargin=1.5em]
%     \item $x^{\mathrm{obj}}(t)$: object position at time $t$.
%     \item $x^{\mathrm{goal}}$: target goal position.
%     \item $d_{\mathrm{goal}}(t)$: object--goal residual distance.
%     \item $L_{\mathrm{obj}}$: characteristic object size.
%     \item $d_{\mathrm{goal}}^{\max}$: normalization scale, set to $L_{\mathrm{obj}}$.
%     \item $T_k$: number of steps in the $k$-th stage.
% \end{itemize}

% -----------------------------------
% \paragraph{Stage Completion Rule.}
% A stage is considered completed when its potential exceeds a threshold 
% for several consecutive steps:
% \[
% \Phi_{\text{stage}_k}(s_t)>\theta_k
% \quad\text{for }3{\text{--}}5\text{ steps}.
% \]

\subsection{Notation Summary}

Table~\ref{tab:notation} lists all variables used in the stage-wise cost and
potential definitions across all tasks.

\begin{table}[!h]
\centering
\begin{tabular}{ll}
\toprule
Symbol & Meaning \\
\midrule
$x^{\mathrm{ee}}(t)$ & End-effector (TCP) position at time $t$. \\
$x^{\mathrm{obj}}(t)$ & Object center or grasp-point position. \\
$x^{\mathrm{goal}}$ & Target goal position of the object. \\
$x^{\mathrm{obj}}_{\mathrm{init}}$ & Object position at episode start. \\

$d_{\mathrm{reach}}(t)$ & TCP–object distance during Reach. \\
$d_{\mathrm{pose}}(t)$ & TCP–object alignment error during Grasp. \\
$d_{\mathrm{trans}}(t)$ & Object–goal residual distance for Transport. \\
$d_{\mathrm{place}}(t)$ & Near-goal deviation during Place. \\
$d_{\mathrm{pp}}(t)$ & Residual distance for Push\&Pull. \\
$d_{\mathrm{lift}}(t)$ & Vertical height error. \\
$d_{\mathrm{upright}}(t)$ & Geodesic rotation distance on $\mathrm{SO}(3)$. \\
$d^{\max}_{\cdot}$ & Normalization scales (stage-dependent). \\
$L_{\mathrm{obj}}$ & Characteristic object size (e.g., cube side length). \\
$T_k$ & Number of steps in stage $\tau^{(k)}$. \\
$R_{\mathrm{obj}}(t)$ & Object orientation at time $t$. \\
$R_{\mathrm{upright}}$ & Target upright orientation. \\
\bottomrule
\end{tabular}
\caption{Summary of notation used across all stage definitions.}
\label{tab:notation}
\end{table}

% \paragraph{Remark.}
% All potentials are non-negative by design, with values in $[0,1]$. 
% Tolerance parameters such as $d_\ast$ or $h_\text{lift}$ define acceptable ranges, 
% ensuring robustness to small deviations and avoiding brittle binary signals.

\section{Broader Impacts}
Our framework aims to improve safety and reliability of robot manipulation by aligning policies with human-preferred, semantically correct behaviors. 
Potential risks include overfitting to biased preferences and misuse in unsafe settings; we discuss mitigations in the main text and Appendix.

\end{document}